\documentclass[10pt,twocolumn,letterpaper]{article}

\usepackage[pagenumbers]{wacv} %

\usepackage[utf8]{inputenc}
\usepackage{graphicx}
\usepackage{amsmath}
\usepackage{amssymb}
\usepackage{booktabs}
\usepackage{subcaption}
\usepackage{float}
\usepackage{rotating}
\usepackage[T1]{fontenc}    

\usepackage[pagebackref,breaklinks,colorlinks]{hyperref}
\newcommand{\ours}{\textbf{SimuScope}}

\usepackage[capitalize]{cleveref}
\crefname{section}{Sec.}{Secs.}
\Crefname{section}{Section}{Sections}
\Crefname{table}{Table}{Tables}
\crefname{table}{Tab.}{Tabs.}

\def\wacvPaperID{1448} %
\def\confName{WACV}
\def\confYear{2025}

\begin{document}

\title{SimuScope: Realistic Endoscopic Synthetic Dataset Generation through Surgical Simulation and Diffusion Models}

\author{
Sabina Martyniak\textsuperscript{1} \quad
Joanna Kaleta\textsuperscript{1,2} \quad
Diego Dall'Alba\textsuperscript{3} \quad
Michał Naskręt\textsuperscript{1} \quad \\
Szymon Płotka\textsuperscript{1,4} \quad
Przemysław Korzeniowski\textsuperscript{1} \quad
\\[.5em]
Sano Centre for Computational Medicine, Poland\textsuperscript{1} \quad
Warsaw University of Technology, Poland\textsuperscript{2} \quad \\
University of Verona, Italy\textsuperscript{3} \quad 
University of Warsaw, Poland\textsuperscript{4} \quad
}

\maketitle

\begin{abstract}
Computer-assisted surgical (CAS) systems enhance surgical execution and outcomes by providing advanced support to surgeons. These systems often rely on deep learning models trained on complex, challenging-to-annotate data. While synthetic data generation can address these challenges, enhancing the realism of such data is crucial. This work introduces a multi-stage pipeline for generating realistic synthetic data, featuring a fully-fledged surgical simulator that automatically produces all necessary annotations for modern CAS systems. This simulator generates a wide set of annotations that surpass those available in public synthetic datasets. Additionally, it offers a more complex and realistic simulation of surgical interactions, including the dynamics between surgical instruments and deformable anatomical environments, outperforming existing approaches. To further bridge the visual gap between synthetic and real data, we propose a lightweight and flexible image-to-image translation method based on Stable Diffusion (SD) and Low-Rank Adaptation (LoRA). This method leverages a limited amount of annotated data, enables efficient training, and maintains the integrity of annotations generated by our simulator. The proposed pipeline is experimentally validated and can translate synthetic images into images with real-world characteristics, which can generalize to real-world context, thereby improving both training and CAS guidance. The code and the dataset are available at \url{https://github.com/SanoScience/SimuScope}. 
\end{abstract}

\section{Introduction}

Computer Assisted Surgery (CAS) is a rapidly evolving field that aims to enhance surgical procedures by providing advanced technological support to surgeons \cite{cai4cai}. By integrating sophisticated computational tools, CAS systems can improve the precision, safety, and outcomes of surgeries \cite{zaffino2020review,schneider2021performance,prevost2020efficiency,plotka2025real}. One of the most extensively studied procedures within this domain is cholecystectomy, a surgical procedure for the removal of the gallbladder \cite{tokuyasu2021development}. Despite its prevalence in surgery, cholecystectomy can lead to serious complications such as bile duct injury, which underscores the need for improved surgical support systems \cite{tokuyasu2021development}.
\begin{figure}[t!]
    \centering
    \captionsetup[subfigure]{labelformat=empty}
    \vspace*{0cm}
    \begin{subfigure}{\linewidth}
        \includegraphics[width=\linewidth]{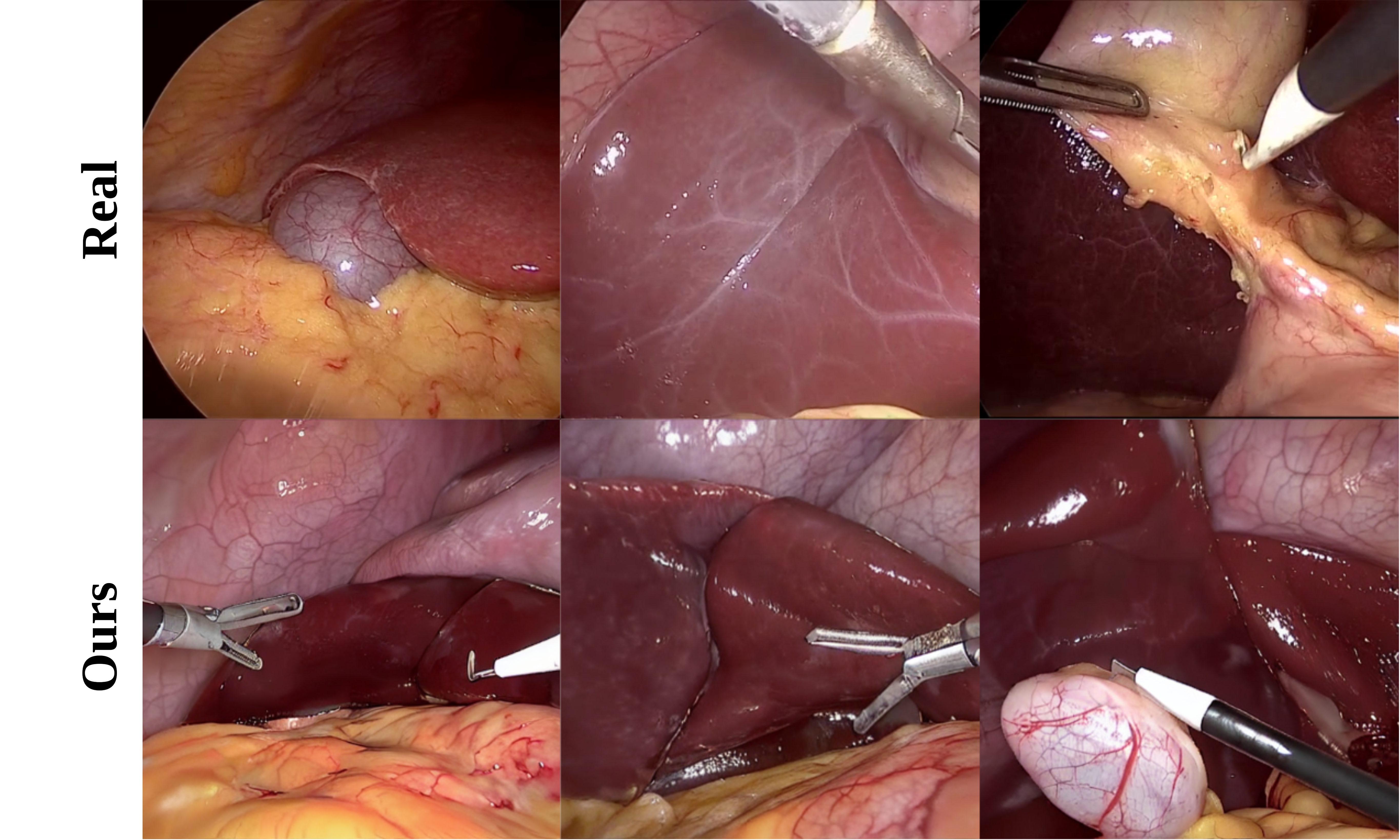}
    \end{subfigure}
    \caption{Comparison of real images and images generated by \ours{}. The real images used in this comparison are sourced from the CholecT45 dataset \cite{cholect45}, which comprises authentic surgical footage. Both the real and generated images exhibit comparable coloration and textural details, posing a challenge in distinguishing between them at a glance. This similarity underscores the fidelity and realism achieved by SimuScope in simulating surgical scenarios.}
    \label{fig:ourtoreal}
\end{figure}
CAS systems can leverage deep learning (DL) methods to perform various supporting tasks. These include the segmentation of anatomical structures and surgical instruments \cite{lou2023min,bhattarai2023histogram}, as well as the temporal modeling \cite{jin2021temporal} of the procedure at different levels of detail. These DL systems' effectiveness heavily depends on large volumes of annotated data. However, obtaining such data is challenging due to the need for detailed and time-consuming annotations, which can only be performed by expert personnel. This requirement poses a significant limitation on the development and scalability of CAS systems.

To address this limitation, synthetic data generated through virtual simulators is proposed. These simulators automatically generate the necessary annotations, reducing the dependency on manual annotation. However, existing simulators often lack realism and provide limited interactions between surgical tools and tissues. Additionally, the annotations generated by these simulators are typically restricted to specific tasks, such as semantic segmentation of the scene.

In this work, we propose an advanced virtual simulator that overcomes these limitations. Our simulator offers a highly realistic biomechanical simulation of tissues and rich interactions, such as tissue grasping, tearing, cutting, thermocoagulation, and vessel clipping with various types of surgical instruments, including both laparoscopic and robotic ones. This enhanced realism is crucial for training DL models to support the execution of surgical procedures effectively. Furthermore, our simulator generates an extensive set of annotations, including pixel-level semantic segmentation, depth and normal maps, optical flow, as well as temporal and high-level text information, such as surgical action-triplets \cite{cholect45} and instrument poses in both 2D and 3D. These detailed annotations provide a more comprehensive dataset for training DL models, improving their performance and reliability.

To further enhance the visual realism of the main endoscopic RGB image rendered by our simulator, we propose an image-to-image (img2img) translation approach. It is based on SD combined with ControlNet, which allows us to preserve the automatically generated annotations while significantly improving the visual quality of the rendered images. Unlike traditional methods based on Generative Adversarial Networks (GANs) \cite{goodfellow2020generative}, our approach is lightweight and efficient, utilizing multiple LoRA modules \cite{LoRa_2021}. This enables us to achieve high-quality results with limited annotated data and efficient training processes. The main contributions of our work are as follows:

\begin{enumerate} 

    \item We introduce a high-fidelity virtual simulator that provides a realistic biomechanical simulation of tissues and a rich set of interactions between surgical instruments and tissues, such as grasping, tearing, cutting, clipping, and thermocoagulation. To the best of our knowledge, this is the first work that supports such a wide range of surgical interactions, which is indispensable for training effective DL models. 
    \item We propose a novel image-to-image translation approach that improves the visual realism of synthetic data generated by the simulator while preserving the integrity of all annotations. This approach is both lightweight and efficient, making it a practical solution for enhancing synthetic data quality. 
    \item The simulator generates a comprehensive set of annotations, including detailed temporal information, which significantly enhances the training dataset for DL models. 

\end{enumerate}

\section{Related work}

Recently, several approaches have been proposed for generating synthetic data with realistic characteristics, either for specific surgical procedures or general anatomical structures (e.g., \cite{spinabifida, saramis}). The combination of synthetic images and real segmentation maps is used to train GANs for image analysis and surgical applications, as seen in \cite{Endo-Sim2Real, LC-GAN, mutuallyGAN, Oda_2019}.

While GAN-based approaches show potential, they have limitations, such as early convergence of discriminators and instability of adversarial training, leading to mode collapse and reduced diversity in generated data \cite{longlivegan}. Diffusion models (DMs) \cite{ddpm} emerge as a promising alternative, surpassing GANs in computer vision tasks.

Diffusion models are widely adopted in the medical domain with several applications. One is Image-to-Image Translation, such as CT-to-MRI translation \cite{Kim_2024_WACV, diffusion_adversarial}. To address the lengthy training times required for diffusion models, some works focus on zero-shot approaches \cite{zero_shot_frequency, wang_diff_zero_shot, zero_shot_transl}. Diffusion models provide powerful representations useful for image understanding tasks, including segmentation \cite{heidari2023hiformer, kim2023diffusion_vessel, segm_ambiguos, bieder2023memory, medsegdiff}, classification \cite{yang_diffmic}, and anomaly detection \cite{anomaly_diff, anomaly_diff_simplex}. Other applications include image reconstruction \cite{dolce} and image registration \cite{kim2022diffusemorph}.

Data generation is one of the primary objectives of diffusion models, which are widely applied in various styles. Generated data targets different aspects, such as temporal consistency \cite{temporal_diff}, data debiasing \cite{skorupko2024debiasingcardiacimagingcontrolled}, and multimodal generation \cite{jiang2023coladiff, zhan2024medm2gunifyingmedicalmultimodal}.

\begin{figure*}[t!]
    \centering
    \vspace*{0cm}
    \begin{subfigure}{1\linewidth}
        \includegraphics[width=\linewidth]{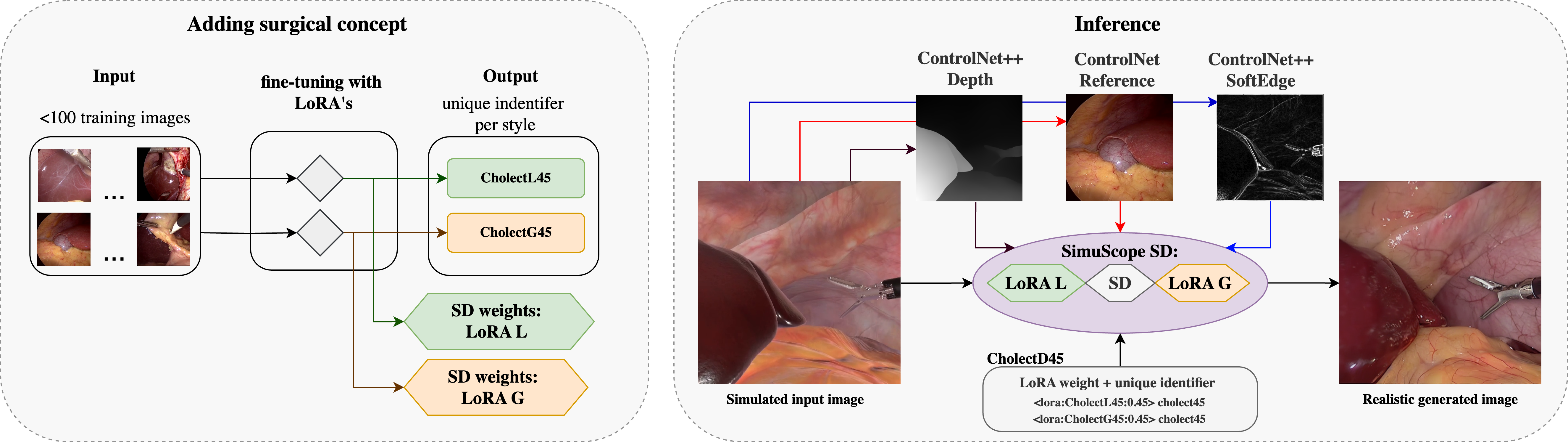}
    \end{subfigure}
    \caption{An overview of SimuScope fine-tuning and inference stage. SD model undergoes fine-tuning using LoRA, a framework that associates a unique LoRA identifier and weight with the newly integrated cholect45 style. During inference, the enhanced SimuScope leverages three ControlNet/ControlNet++ models for comprehensive conditioning. The raw input sample, along with the prompts 'lora:CholectL45:0.45 cholect45' and 'lora:CholectG45:0.45 cholect45,' is fed into SimuScope. The SoftEdge ControlNet++  processes edges predicted by HED, the Depth ControlNet++ handles depth detected by MiDaS, and the Reference ControlNet utilizes additional real input sample as a reference. This multi-model integration enhances SimuScope's capability to generate realistic surgical simulations enriched with detailed texture, edge, depth, and reference data.}
    \label{fig:ConceptInference}
\end{figure*}

Although multiple generative works exist in the medical and surgical fields, a gap remains in surgical data generation, particularly for fully labeled simulator-based data with accurate and detailed instrument-tissue interactions \cite{Kaleta2024,kim2022diffusemorph}. Physically accurate data generation with rich labeling, such as depth, normals, and triplets, is potentially useful for robotic tasks. The closest generative works focusing on laparoscopic cholecystectomy include two studies \cite{img2img_2019, vid2vid_2021}—one on photo collection and one for videos with temporal consistency, both based on GANs. These works require large datasets and use simple simulators with long training times. Although they release a large open-source simulation dataset, it has limitations, including a very simple simulator that lacks tool-tissue interactions and requires long training times.

In \cite{Venkatesh2024}, semantic consistency in unpaired image translation to generate data for surgical applications is investigated, providing insightful analysis and proposing techniques to enhance GAN performance. Based on simulation data, \cite{Kaleta2024} generates large fully labeled realistic endoscopic images, addressing the minimal data requirement. However, this work also relies on a very simple simulator lacking tool-tissue interactions.

\section{Methods}
Our work involves a three-step process as follows. We generate synthetic images of the cholecystectomy procedure from our simulator. We fine-tune the Stable Diffusion (SD) model on a small subset of real images using two LoRAs. Finally, in the sim-to-real phase, we perform inference using the fine-tuned SD model in image-to-image mode, enhanced with three versions of the ControlNet architecture, to ensure consistency between the labels and the generated images. An overview of our method is presented in Figure~\ref{fig:ConceptInference}.

\subsection{Simulation system}

The simulation software runs on a desktop computer equipped with an Intel Core i9-12900K, 64GB RAM, and NVIDIA RTX 3080 GPU. We use two 3D Systems Touch devices for haptic input. The development environment includes the Unity3D game engine integrated with our custom-built surgical simulation framework.

A multi-threaded implementation of the Extended Position-Based Dynamics (XPBD) physics solver, programmed in C/C++, is employed for real-time soft tissue simulation. Recent research (e.g., \cite{SmallSteps, macklin2021constraint, RigidBodies}) establishes XPBD as a competitive alternative to more complex methods for simulating non-linear tissue behavior due to its accuracy, stability, speed, and ease of implementation. The local nature of the non-linear Gauss-Seidel solver within XPBD avoids the limitations associated with global, matrix-based solvers. This enables efficient and accurate implicit simulation of arbitrary elastic and dissipative energy potentials, leading to robust handling of equality and inequality constraints. Additionally, XPBD provides constraint force estimates crucial for accurate haptic feedback calculations in surgical simulation.

To create a virtual model of the liver, gallbladder with cystic duct and artery, and surrounding tissues, we use volumetric tetrahedral meshes consisting of about 50,000 elements in total. A Neo-Hookean constitutive model \cite{macklin2021constraint} is employed to simulate the near incompressibility of soft tissue. The model can conserve volume more than co-rotational finite element or Saint-Venant-Kirchhoff models and can recover from inverted element configurations (i.e., flipped tetrahedrons). The following equation presents the energy-based formulation of the Neo-Hookean model adopted in this simulator:

\begin{equation}
\begin{split}
\Psi_{Neo} &= \Psi_{H} + \Psi_{D} =\\
&= \frac{\lambda}{2} \left(\det (F) - 1\right)^{2} + \frac{\mu}{2} \left(\operatorname{tr}(F^{T}F) - 3\right),
\end{split}
\end{equation}

\noindent
where $F$ is a $3 \times 3$ deformation gradient matrix, $\lambda$ and $\mu$ are the Lamé parameters, $\Psi_{H}$ represents the hydrostatic energy component resisting volume changes, and $\Psi_{D}$ signifies the deviatoric energy component resisting distortion.

The proposed simulator leverages XPBD's iterative constrained optimization approach, which only requires the computation of first-order gradients. This eliminates the need for complex calculations involving Hessian matrices, eigenvalue decomposition, and sophisticated linear solvers, which are characteristic of Newton-method-based approaches \cite{macklin2021constraint}.

The compliance parameters governing both hydrostatic and deviatoric constraints are visually tuned to approximate the behavior of real anatomical structures. This tuning process is guided by the feedback of experienced surgeons interacting with the virtual environment. Despite the computational demands, refresh rates up to 1.5 kHz are achieved, allowing for small simulation time steps (0.75–1.0 ms), which are crucial for accurate haptic interaction.

\subsection{Virtual Cholecystectomy Surgery}

\begin{figure}[t!]
    \centering
    \vspace*{0.2cm}
    \begin{subfigure}{0.32\linewidth}
        \includegraphics[width=\linewidth]{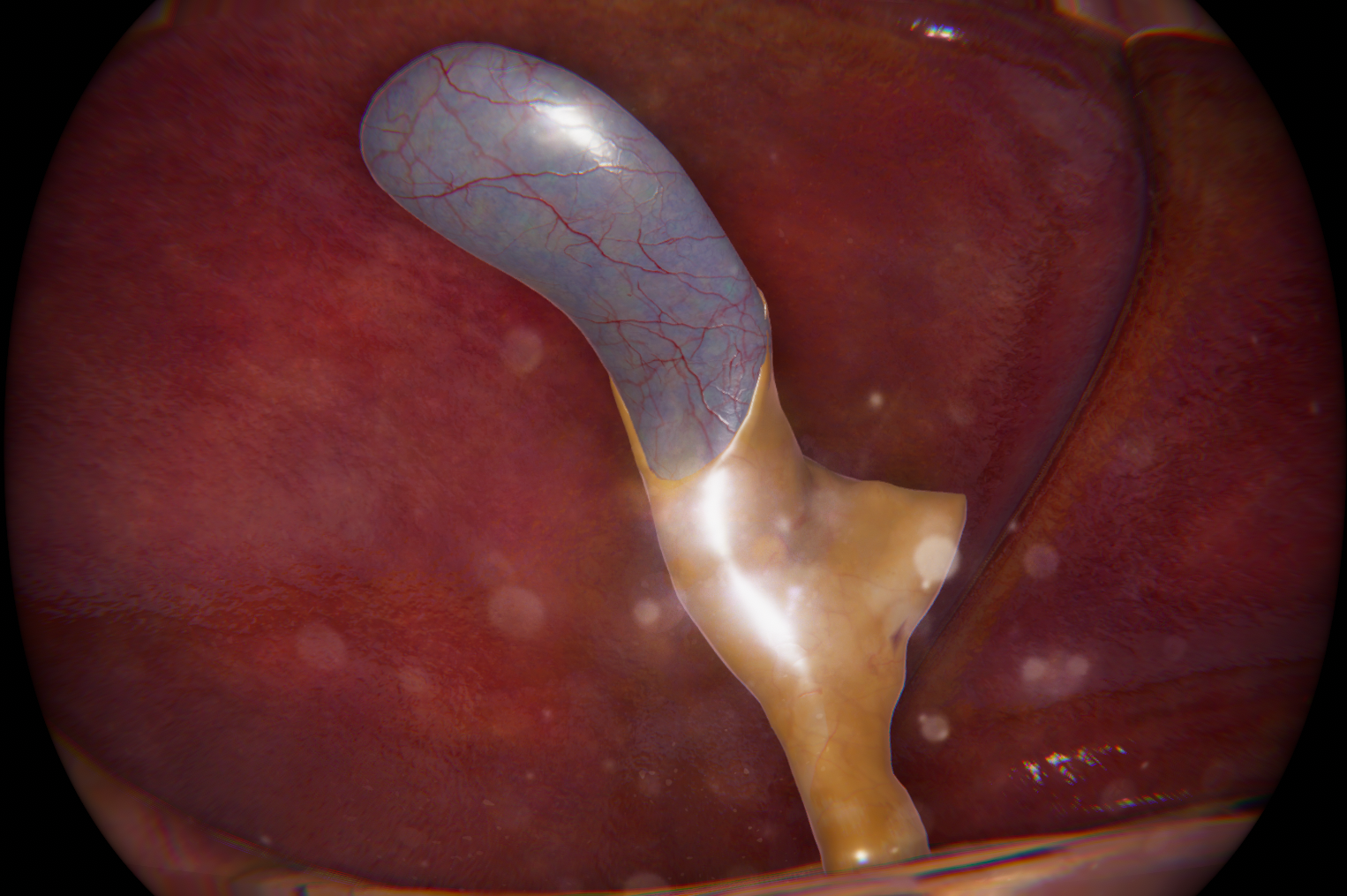}
        \caption{}
    \end{subfigure}
    \begin{subfigure}{0.32\linewidth}
        \includegraphics[width=\linewidth]{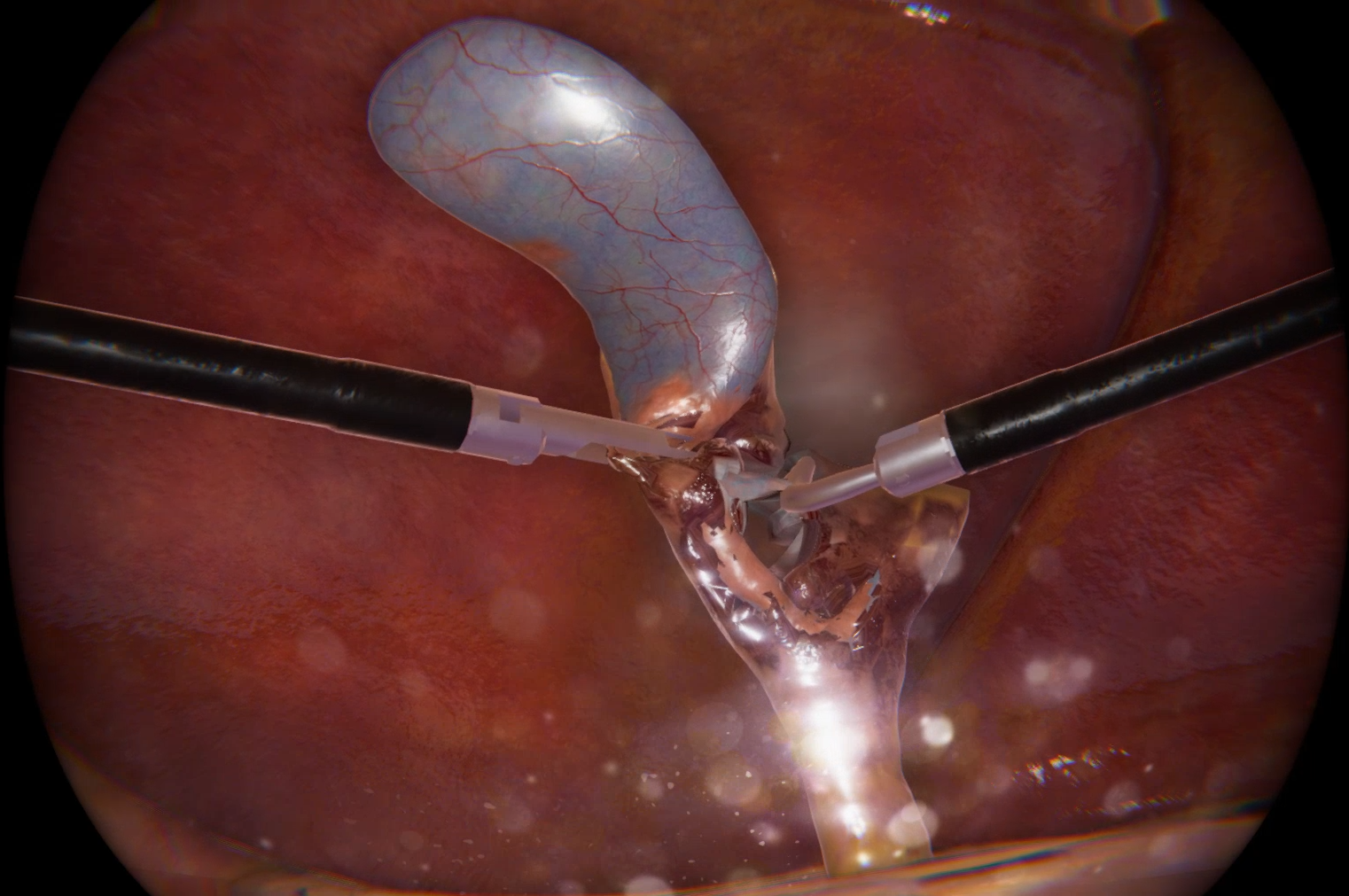}
        \caption{}
    \end{subfigure}
    \begin{subfigure}{0.32\linewidth}
        \includegraphics[width=\linewidth]{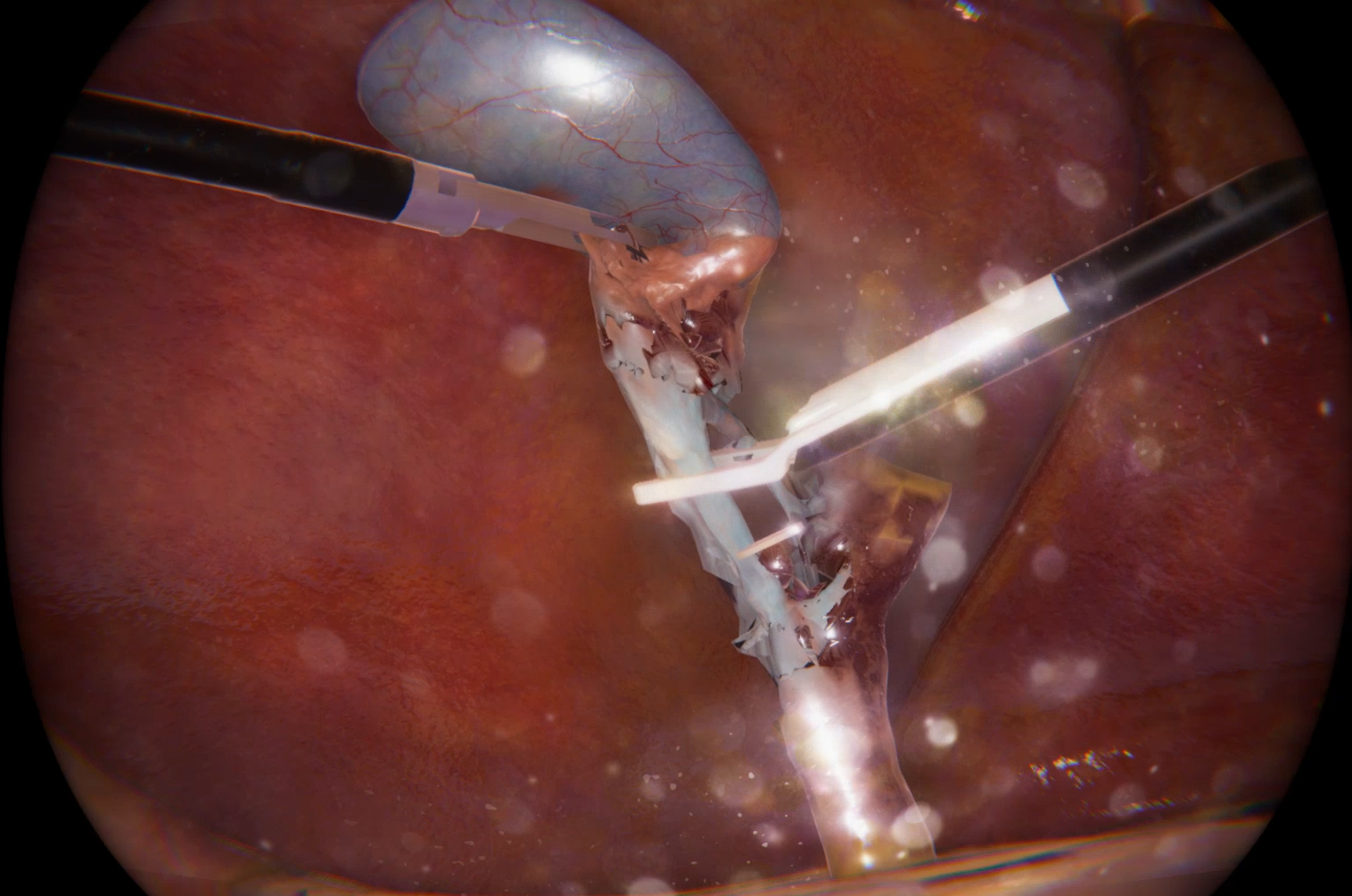}
        \caption{}
    \end{subfigure}
    \begin{subfigure}{0.32\linewidth}
        \includegraphics[width=\linewidth]{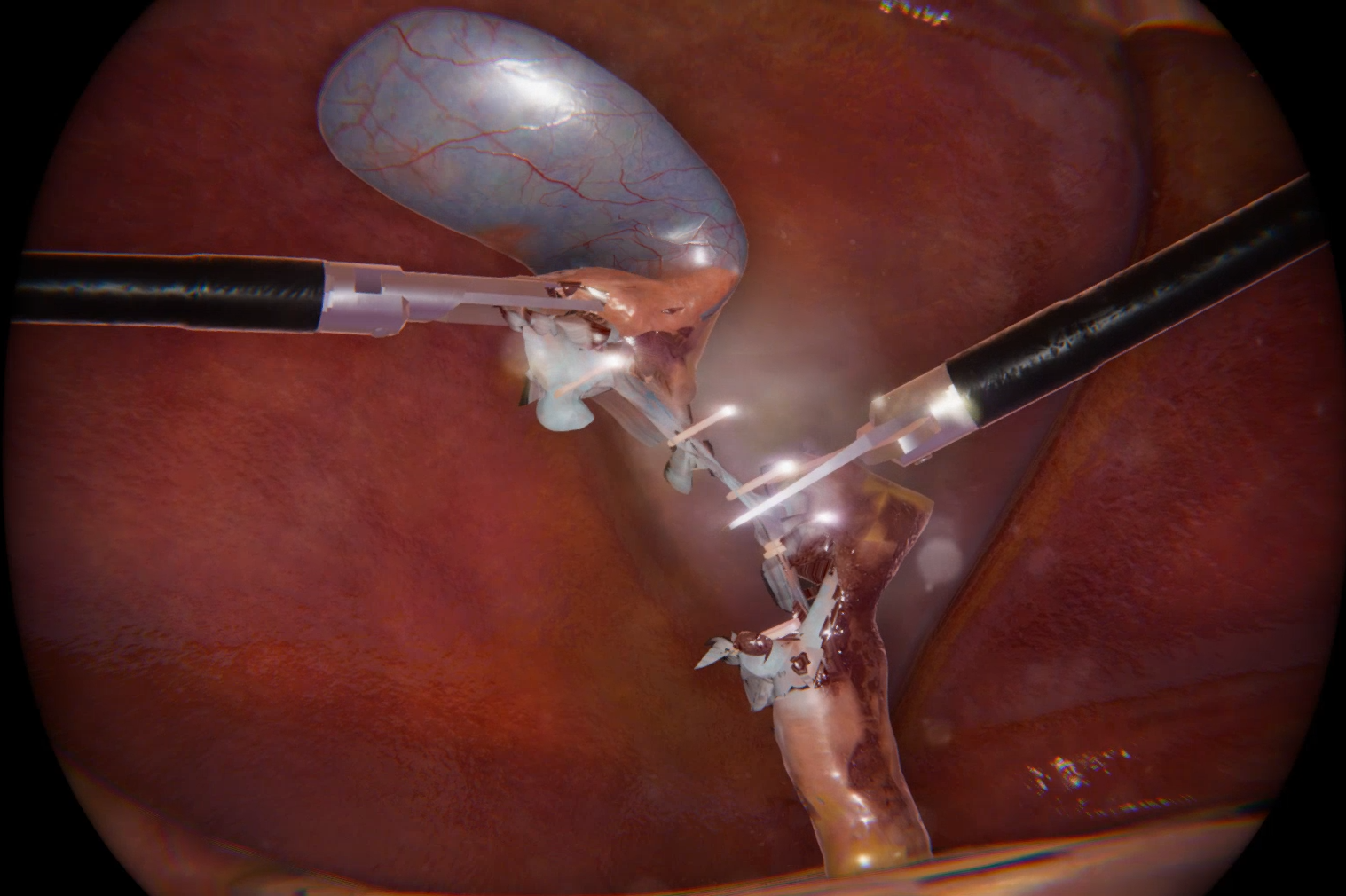}
        \caption{}
    \end{subfigure}
    \begin{subfigure}{0.32\linewidth}
        \includegraphics[width=\linewidth]{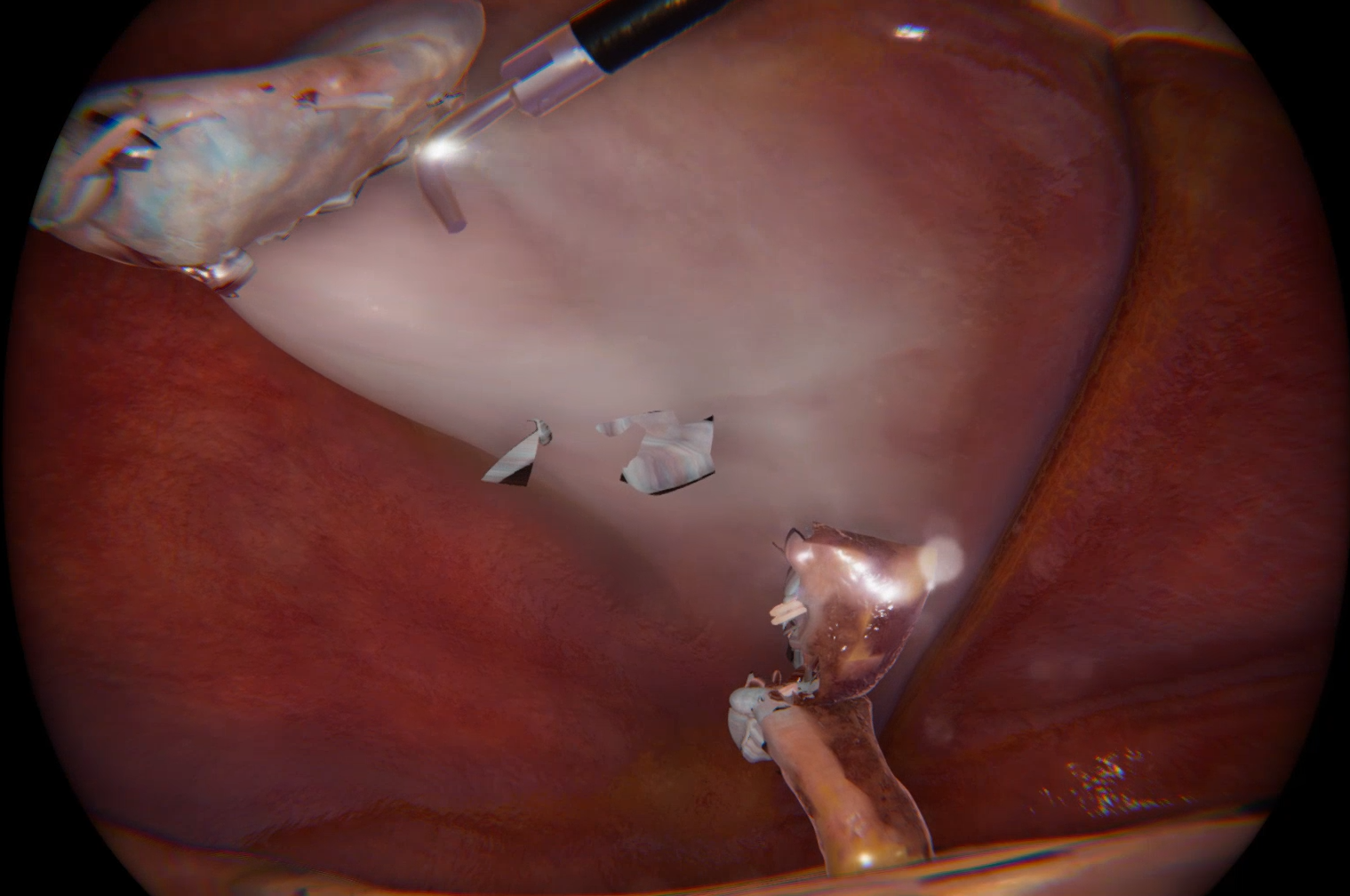}
        \caption{}
    \end{subfigure}
    \begin{subfigure}{0.32\linewidth}
        \includegraphics[width=\linewidth]{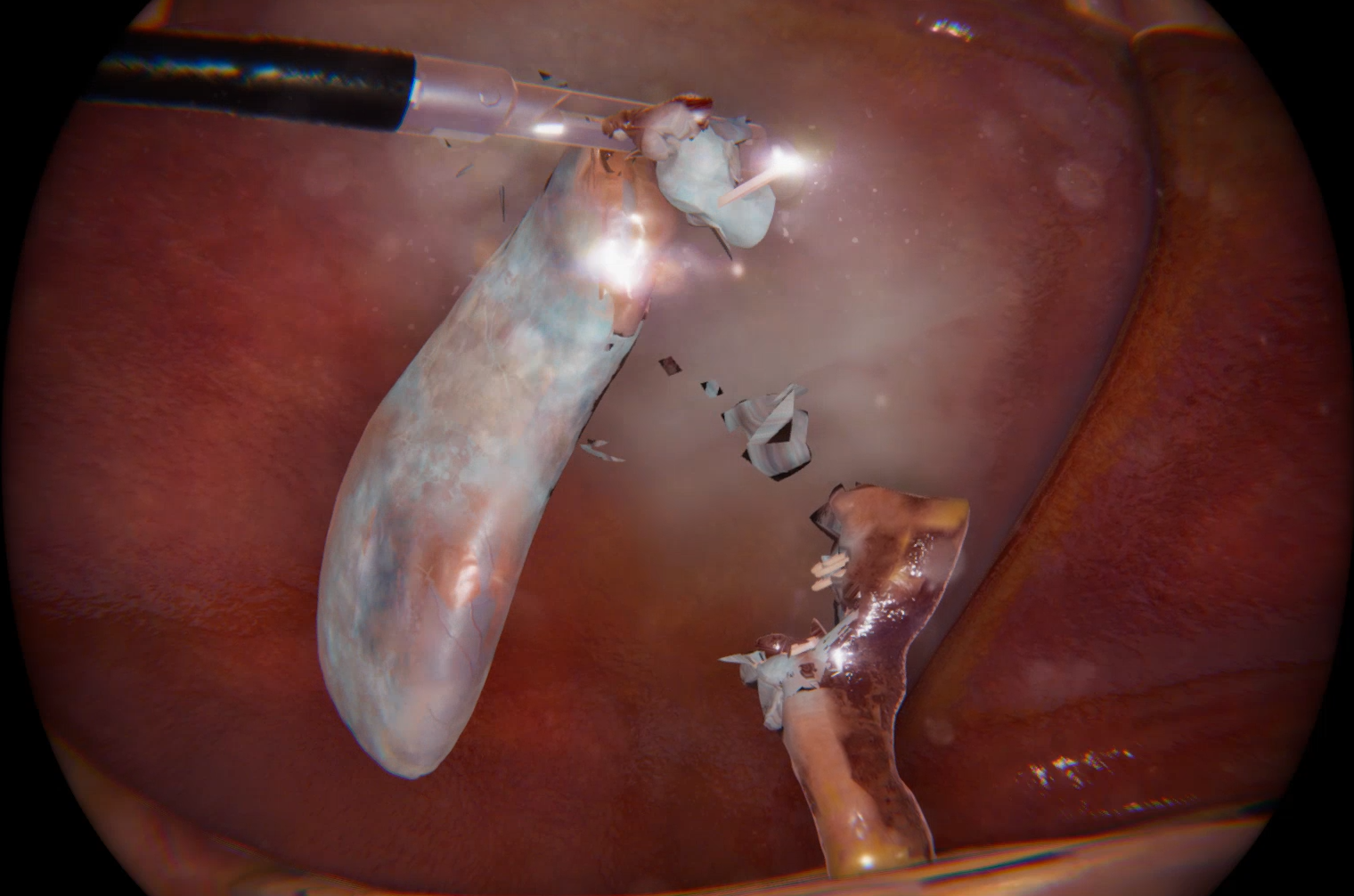}
        \caption{}
    \end{subfigure}
    \caption{An overview of the stages of a virtual cholecystectomy: (a) the start of the procedure, showing the initial setup with surgical instruments inserted into the abdominal cavity; (b) the dissection of Calot's triangle using a grasper and diathermy hook; (c) the clipping of the cystic duct and artery with a clipping tool; (d) the cutting of the cystic duct and artery with scissors; (e) the dissection of the gallbladder from the liver bed using a hook; and (f) the gallbladder fully dissected and ready for removal from the abdominal cavity.}
    \label{fig:vir_lapcholy_steps}
\end{figure}

We chose cholecystectomy for its prevalence in general surgery. The simulation starts with surgical instruments inserted into the abdominal cavity inflated with carbon dioxide gas. Simulation visualizes the liver, gallbladder, cystic duct and artery, which are initially covered with fatty connective tissue (Figure~\ref{fig:vir_lapcholy_steps}(a)). The first stage of the procedure is to use grasper and diathermy hook to dissect the hepatocystic triangle (Figure~\ref{fig:vir_lapcholy_steps}(b)). Next, the operator needs to establish the critical view of safety, which allows to safely clip with the clipping tool (Figure~\ref{fig:vir_lapcholy_steps}(c)) and cut cystic duct and artery with scissors (Figure~\ref{fig:vir_lapcholy_steps}(d)). Finally, the gallbladder is separated from the liver bed using the hook (Figure~\ref{fig:vir_lapcholy_steps}(e) and Figure~\ref{fig:vir_lapcholy_steps}(f)). At this development stage, the simulator does not provide support for irrigation and removal of the gallbladder from the abdominal cavity using a specimen bag. 

\begin{figure}
    \centering
    \includegraphics[width=0.45\textwidth]{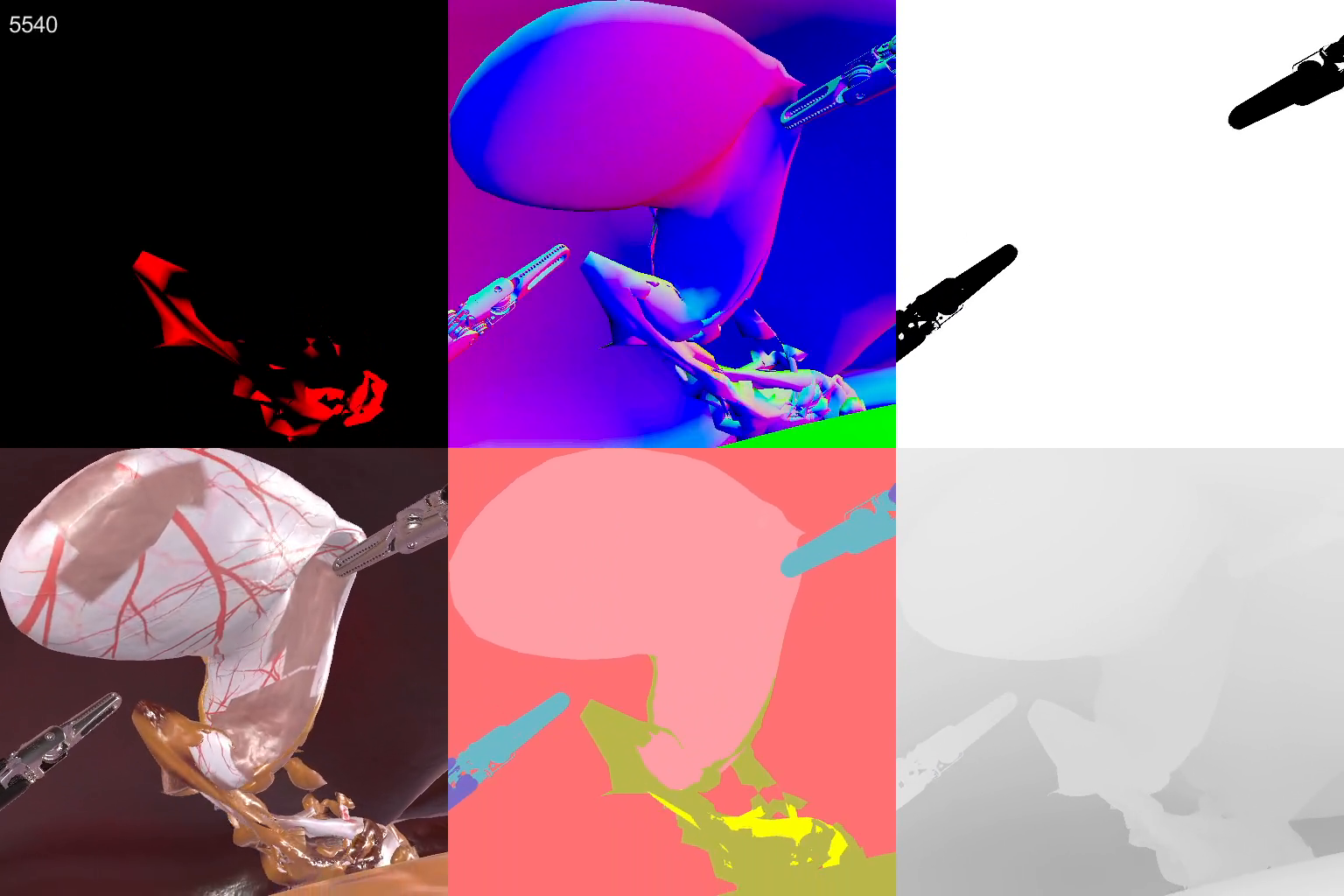}
    \caption{The output from the simulator shows partially dissected Calot's triangle. Top row: blood map - marking the bleeding tissue, normal map - providing detailed surface orientation information to enhance the realism of the simulation, and tools mask - highlighting the specific surgical instruments in use. Bottom row: color - depicting the visual appearance of the surgical scene, segmentation - categorizing different anatomical structures and tools for precise identification, and depth maps - offering information on the distance of objects within the scene to facilitate accurate spatial understanding.}
    \label{fig:simout}
\end{figure}

The simulator, in addition to the endoscopic RGB image, outputs a range of corresponding ground-truth image data, such as depth and normal maps, optical flow, tool masks, semantic segmentation, and procedure-specific masks, including tissue bleeding, damage, and coagulation (Figure~\ref{fig:simout}). It also provides 2D/3D instrument poses and surgical action triplets (e.g., 'grasper retracts gallbladder,' 'hook coagulates cystic duct')~\cite{cholect45}.

\begin{figure*}[h!]
    \centering
    \captionsetup[subfigure]{labelformat=empty}

    \begin{minipage}{0.95\linewidth}
        \centering
        \begin{subfigure}{0.23\linewidth}
            \centering
            \caption{Segmentation}
            \includegraphics[width=\linewidth]{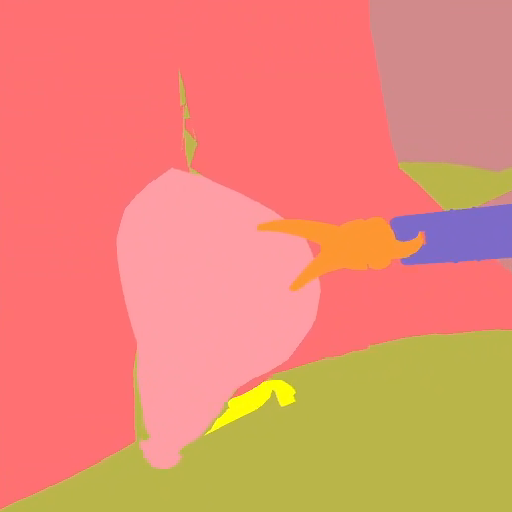}
        \end{subfigure}
        \begin{subfigure}{0.23\linewidth}
            \centering
            \caption{Simulator}
            \includegraphics[width=\linewidth]{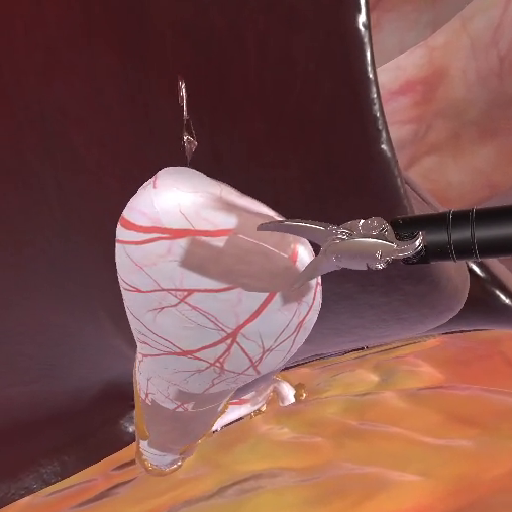}
        \end{subfigure}
        \begin{subfigure}{0.23\linewidth}
            \centering
            \caption{SoftEdge+Depth+Reference}
            \includegraphics[width=\linewidth]{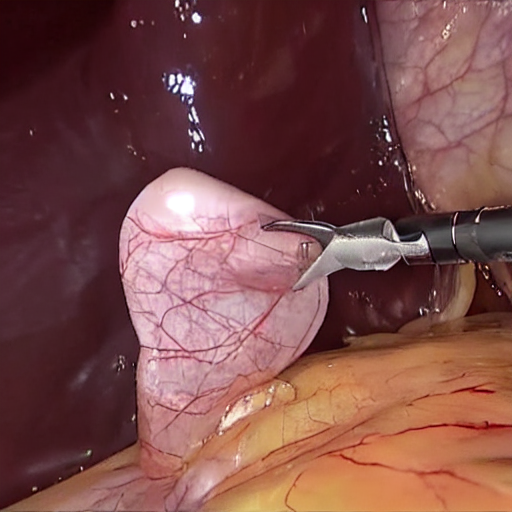}
        \end{subfigure}
        
        \begin{subfigure}{0.2\linewidth}
            \centering
            \caption{No control}
            \includegraphics[width=\linewidth]{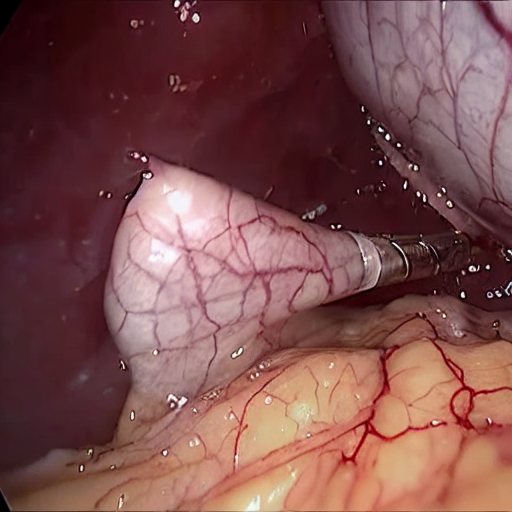}
        \end{subfigure}
        \begin{subfigure}{0.2\linewidth}
            \centering
            \caption{SoftEdge+Reference}
            \includegraphics[width=\linewidth]{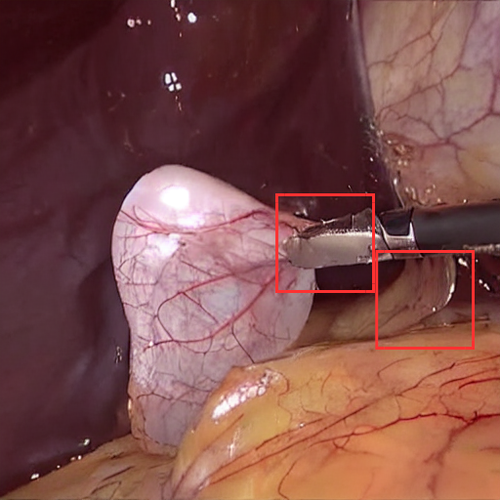}
        \end{subfigure}
        \begin{subfigure}{0.2\linewidth}
            \centering
            \caption{SoftEdge+Depth}
            \includegraphics[width=\linewidth]{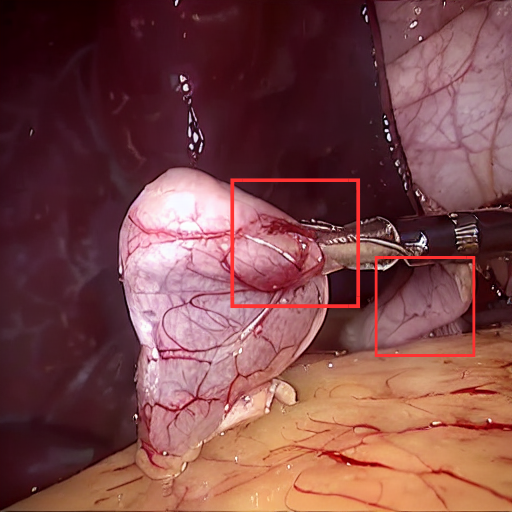}
        \end{subfigure}
        \begin{subfigure}{0.2\linewidth}
            \centering
             \caption{Reference+Depth}
            \includegraphics[width=\linewidth]{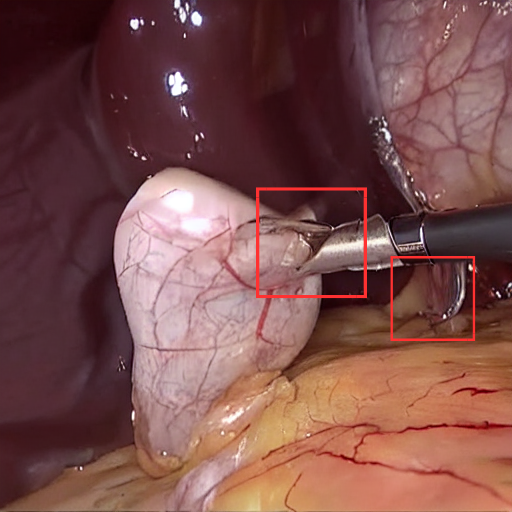}
        \end{subfigure}
    \end{minipage}

        \caption{Visual comparison of sample images generated using different types of controls. Without any control, the overall image consistency is degraded, resulting in inconsistent textures and inaccurate color representation.}
    \label{fig:ControlNet}
\end{figure*}

\subsection{Fine-tuning with LoRAs}
LoRA \cite{LoRa_2021} is an innovative method for efficiently fine-tuning large language models. Instead of retraining all the model's parameters, LoRA introduces trainable rank decomposition matrices into each layer of the model's architecture while keeping the pre-trained model weights frozen. This approach significantly reduces the number of trainable parameters for downstream tasks, leading to a substantial decrease in GPU memory requirements and overall costs, without compromising model performance. LoRA is primarily applied to the attention blocks of Transformers in Large Language Models \cite{vaswani2017attention}, where researchers find that LoRA fine-tuning provides similar quality to full model fine-tuning while being faster and requiring less computational resources.

For instance, fine-tuning an SD model can be achieved using DreamBooth (DB) in LC-SD \cite{dreambooth_2022}, a method that adjusts the entire model to align with a particular concept or style. Although DB yields impressive outcomes \cite{Kaleta2024,kaleta2023lc}, it has a major drawback: the size of the fine-tuned model. Since DB updates the entire model, the resulting checkpoint can be quite large (around 2 to 7 GB) and demands substantial GPU resources for training. In contrast, a LoRA adapter requires significantly less GPU power, yet its inferences are still comparable to those of a DB fine-tuned checkpoint. For comparison, a model fine-tuned using DB occupies about 2GB \cite{Kaleta2024} while LoRA takes up 0.10 GB, which reduces the required space by over 20 times. This allows for a significant reduction in file sizes while maintaining comparable performance, resulting in lower GPU resource requirements in our models.

We fine-tune SD model using two LoRAs: \textit{CholectG45}, which focuses on accurately representing the gallbladder, and \textit{CholectL45}, which specializes in generating a detailed liver model and surrounding tissues. Together, they ensure anatomical consistency and highly detailed textures.

\subsection{Inference with ControlNet}

Realistic tissues are generated from simulation scenes using image-to-image inference with LoRA adapters. By combining \textit{CholectG45} and \textit{CholectL45}, introduced in the previous section, we create a new style called \textit{CholectD45}  (see Figure \ref{fig:ConceptInference}). 

To improve the precision of text-to-image diffusion models, ControlNet \cite{controlNet_2023} introduces image-based conditioning controls. ControlNet++ \cite{ControlNet++} extends this by optimizing alignment between input conditions and generated images using a pre-trained discriminative reward model for cycle consistency, unlike methods relying on latent diffusion denoising. This enhances controllability across various conditions.

\noindent
\textbf{Alternative control tools.} Various methods impose control on diffusion models. Prompt engineering \cite{li2023gligen,yang2023reco,zhang2023controllable} is useful for general tasks but struggles with enforcing constraints in specialized domains like surgery. T2I-Adapter \cite{mou2024t2i}, conceptually most similar to ControlNet, offers composability and generalization but it is not as widely adopted as ControlNet. Cross-attention constraints \cite{chen2024training,ju2023humansd,xie2023boxdiff} and instance-based controllable generation \cite{wang2024instancediffusion,zhou2024migc} provide regulated generation but lack the precision needed for surgical applications. Compared to the mentioned works, ControlNet and ControlNet++ are widely adopted, offering precise control from various inputs applicable to unusual data domains.

\noindent
\textbf{Applied types of control.} In our work, we utilize both ControlNet \cite{controlNet_2023} and ControlNet++\cite{ControlNet++}. SoftEdge and Depth ControlNet++ models play a crucial role in preserving details and improving shapes and boundaries in the output images. The SoftEdge ControlNet++ model focuses on generating natural tissue appearances by controlling detail through soft edges. We use the HED preprocessor \cite{hed} to detect edges from the input sample, which are then fed into the SoftEdge ControlNet++. Depth ControlNet++ leverages depth information obtained from the MiDaS model \cite{midas} to further enhance image realism and structural accuracy. Reference ControlNet \footnote{https://github.com/Mikubill/sd-webui-controlnet/discussions/1236} enhances control and customization in image generation by enabling users to guide outputs based on a reference image. The Reference model establishes a direct link between the reference image and the model's attention layers, allowing focused replication of specific elements from the reference. It serves as a valuable tool for generating images that closely resemble or are inspired by the reference, enabling greater precision and alignment with user intent.
To achieve the desired balance, we utilize stronger Depth and Reference controls in combination with a weaker SoftEdge control. Removing only one control diminishes the quality of the generated images (see Figure \ref{fig:ControlNet}). To generate data on a large scale while maintaining reasonable inference time and acceptable image quality, we limit the denoising steps to 20. The denoising strength and CFG scale were determined experimentally, as in \cite{Kaleta2024},  with values listed in Supplementary Material Table 1.

\section{Experiments and Results}

In this section, we describe the dataset used for training and evaluating SimuScope. Following this, we outline the implementation details and evaluation metrics. Finally, we present our quantitative results.

\begin{table*}[t]\centering
\caption{Quantitative results on \textbf{our} simulator.We compare the raw simulator output with data generated using a concurrent method. We demonstrate the effectiveness of our approach using key metrics including mIoU, FID, KID, CMMD, Density, and Coverage.}
\resizebox{\textwidth}{!}{%
\begin{tabular}{cccccccc}
\toprule
Method & Style & mIoU [\%] ↑ & FID ↓ & KID ↓ & CMMD ↓ & Density ↑ & Coverage ↑ \\
\midrule
N/A & Raw & 60.79 & 202.54 & .1888 & 3.442 & 0.011 & 0.004\\
LC-SD \cite{Kaleta2024}& Mixed styles & 67.42 &\textbf{68.26}& \textbf{.0635} &1.017&0.077& 0.104 \\
\ours & CholectD45 & \textbf{70.65} & 79.80 & .0690 &  \textbf{0.697} & \textbf{0.114} & \textbf{0.117}\\
\bottomrule
\end{tabular}
}
\label{tab:metrics}
\end{table*}

\subsection{Dataset}
To train SimuScope, we utilized two diverse real image sets with varying visual properties. The set for CholectL45 focuses on liver images from endoscopic views while also adapting to other tissues. The image set for CholectG45 emphasizes gallbladder images, ensuring diversity while incorporating other tissues and instruments. Both LoRAs were fine-tuned on fewer than 85 manually selected training images sourced from CholecT45 \cite{cholect45}. These training images were carefully chosen to represent different stages of the procedure and include a variety of tissues and instruments to ensure comprehensive coverage.

For inference, we created a dataset using video footage from a simulator, which included DaVinci instruments (see Figure 3 in Supplementary Material). The simulator output depicted partially dissected Calot’s triangle and provided data on color, segmentation, depth maps, optical flow, normal maps, and tool masks (see Figure \ref{fig:simout}).

\subsection{Implementation details}

For training the diffusion model, we utilize the implementation from \cite{controlNet_2023}. We perform all experiments using a single NVIDIA A100 80GB GPU. To maximize the benefits of the pre-trained model, we scale the training samples to a resolution of 512 $\times$ 512 pixels, matching the final pretraining resolution of Stable Diffusion 1.5. During the training phase, we only fine-tune the ControlNet branch of the model. This configuration allows us to train the model with a batch size of 1 through gradient accumulation. The model is trained with a learning rate of 1.2 $\times$ 10$^{-3}$ for 20 epochs, with a maximum timestep of 1,000 and a text encoder learning rate of 1.2 $\times$ 10$^{-3}$. We minimize the Mean Squared Error (MSE) loss function with the Adafactor \cite{shazeer2018adafactor} optimizer. 

\subsection{Evaluation metrics}

For evaluation, we employ key metrics to assess the efficacy and fidelity of generated data in computer vision applications. The mean Intersection over Union (mIoU) metric measures the semantic segmentation accuracy by evaluating the overlap between predicted segmentation masks and ground truth annotations.

The Fréchet Inception Distance (FID) \cite{heusel2017gans} quantifies the distance between feature distributions of real and generated images:
\begin{equation}
\text{FID} = \| \mu_X - \mu_Y \|_2^2 + \text{Tr}(\Sigma_X + \Sigma_Y - 2(\Sigma_X \Sigma_Y)^{1/2}),
\end{equation}
\noindent
where \( (\mu_X, \Sigma_X) \) and \( (\mu_Y, \Sigma_Y) \) represent the mean and covariance matrix of the feature distributions of real and generated images, respectively.

The Kernel Inception Distance (KID) \cite{binkowski2018demystifying} evaluates the distributional similarity between real and generated datasets using kernel embeddings:
\begin{equation}
\text{KID} = \| \mu_X - \mu_Y \|_2^2 + \text{Tr}(\Sigma_X + \Sigma_Y - 2(\Sigma_X \Sigma_Y)^{1/2}),
\end{equation}
\noindent
where \( \mu_X \), \( \Sigma_X \) and \( \mu_Y \), \( \Sigma_Y \) denote the mean and covariance matrix of kernel embeddings for real and generated data distributions. Additionally, to assess the quality of the image based on features extracted from the simulator, we use CMMD \cite{cmmd2024} to calculate the result. It is suitable for smaller datasets because it is unbiased, which we cannot claim in the case of FID. To assess sample fidelity, we use the Density and Coverage \cite{D&C} metrics, both of which are derived from nearest neighbors in the representation space. Density quantifies how many neighborhood spheres of real samples encompass a given sample. For measuring sample diversity, we use Coverage, which also relies on the nearest neighbors in the representation space. Density and Coverage indicate the perceptual quality and diversity of the generated images, respectively:

\begin{equation}
\begin{split}
\left( \{x_i^g\}_{i=1}^n, \{x_j^r\}_{j=1}^m \right) = & \, \\ = \frac{1}{kn} \sum_{i=1}^n \sum_{j=1}^m
& \mathbf{1} \left( x_i^g \in B \left( x_j^r, \text{NND}_k \left( x_j^r \right) \right) \right),
\end{split}
\end{equation}

\begin{equation}
\begin{split}
\left( \{x_i^g\}_{i=1}^n, \{x_j^r\}_{j=1}^m \right) = & \, \\ = \frac{1}{m} \sum_{j=1}^m \max_{i=1,\dots,n}
& \mathbf{1} \left( x_i^g \in B \left( x_j^r, \text{NND}_k \left( x_j^r \right) \right) \right).
\end{split}
\end{equation}

\noindent
where \(\mathbf{1}(\cdot)\) indicate the indicator function, \(S(\{x_j^r\}_{j=1}^m) = \bigcup_{j=1}^m B(x_j^r, \text{NND}_k(x_j^r))\), where \(B(x, r)\) defines  a Euclidean ball centered at \(x\) with radius \(r\), and \(\text{NND}_k(x_j^r)\) is the distance between \(x_j^r\) and its \(k\)-th nearest neighbour in \(\{x_j^r\}_{j=1}^m\), excluding itself \cite{D&C}.

\subsection{Quantitative results}

\begin{table}[h]
\caption{Ablation study results showing mIoU (\%) for different combinations of SoftEdge, Reference, and Depth controls.}
\centering
\begin{tabular}{cccc}
    \toprule
    \textbf{SoftEdge} & \textbf{Reference} & \textbf{Depth} & \textbf{mIoU [\%]} \\
    \midrule
    \textcolor{red}{$\times$} & \textcolor{red}{$\times$} & \textcolor{red}{$\times$} & 41.00 \\
    \textcolor{green}{\checkmark} & \textcolor{green}{\checkmark} & \textcolor{red}{$\times$} & 64.39 \\
    \textcolor{green}{\checkmark} & \textcolor{red}{$\times$} & \textcolor{green}{\checkmark} & 54.37 \\
    \textcolor{red}{$\times$} & \textcolor{green}{\checkmark} & \textcolor{green}{\checkmark} & 58.29 \\
    \textcolor{green}{\checkmark} & \textcolor{green}{\checkmark} & \textcolor{green}{\checkmark} & \textbf{70.65} \\
    \bottomrule
\end{tabular}
\label{tab:ablation_results}
\end{table}

We evaluate the efficacy of SimuScope in generating realistic surgical images by comparing them against the raw simulator output using key metrics: mIoU, FID, and KID. Our results, as shown in Table \ref{tab:metrics}, demonstrate substantial improvements over the baseline simulator. Specifically, employing \textit{SimuScope} with style CholectD45 yields a notable increase in mIoU from 60.79\% to 70.65\%, indicating enhanced semantic segmentation accuracy. Moreover, the FID decreases significantly from 202.54 to 79.80, the KID also showed improvement from 0.1888 to 0.0690, the CMMD decrease from 1.017 to 0.697, the Density increased from 0.077 to 0.114 and the Coverage from 0.104 to 0.117. These metrics underscore the capability of SimuScope to produce surgical images that closely resemble real-world scenarios, showcasing its potential for advancing computer-assisted surgery through enhanced training and simulation capabilities.

\begin{figure*}[t!]
    \centering
    \captionsetup[subfigure]{labelformat=empty}
    \begin{minipage}[c]{0.05\linewidth}
        \centering
        \rotatebox{90}{\textbf{Simulator}}
    \end{minipage}%
    \begin{minipage}{0.95\linewidth}
        \centering
        \begin{subfigure}{0.15\linewidth}
            \centering
            \includegraphics[width=\linewidth]{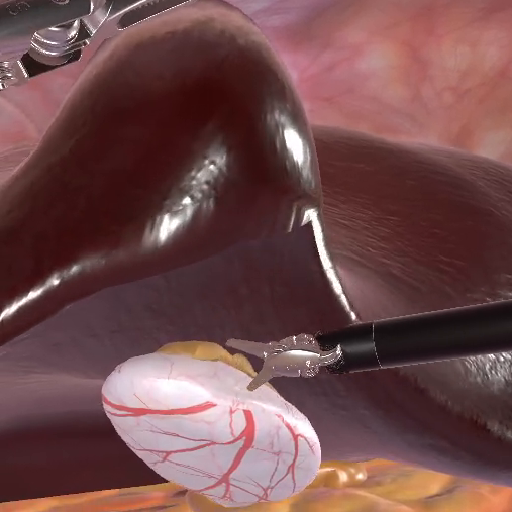}
            \caption{}
        \end{subfigure}
         \begin{subfigure}{0.15\linewidth}
            \centering
            \includegraphics[width=\linewidth]{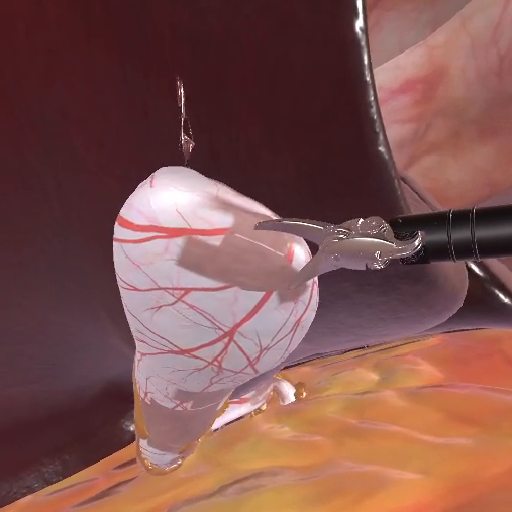}
            \caption{}
        \end{subfigure}
        \begin{subfigure}{0.15\linewidth}
            \centering
            \includegraphics[width=\linewidth]{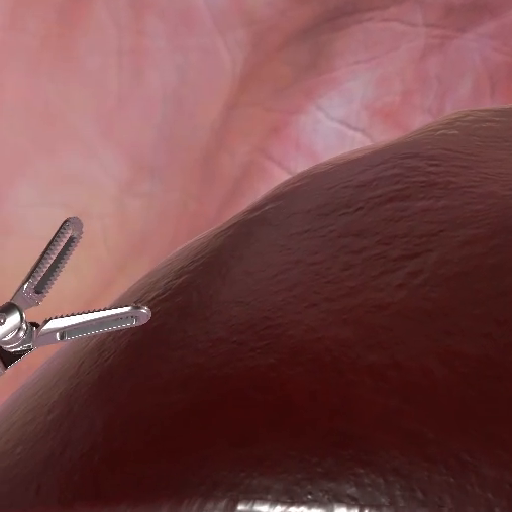}
            \caption{}
        \end{subfigure}
        \begin{subfigure}{0.15\linewidth}
            \centering
            \includegraphics[width=\linewidth]{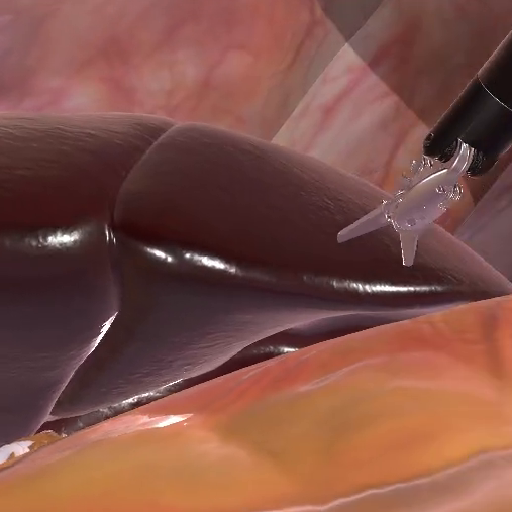}
            \caption{}
        \end{subfigure}
        \begin{subfigure}{0.15\linewidth}
            \centering
            \includegraphics[width=\linewidth]{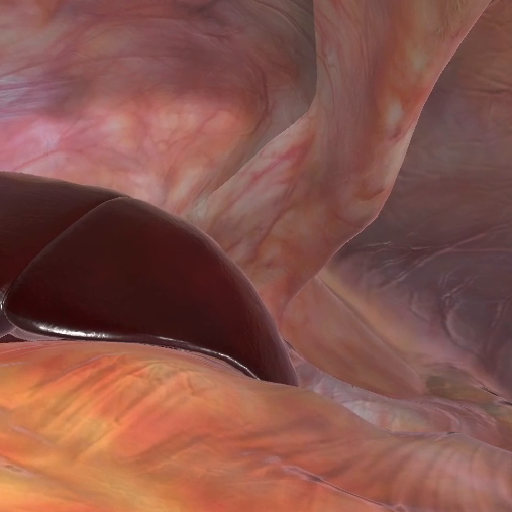}
            \caption{}
        \end{subfigure}
        \begin{subfigure}{0.15\linewidth}
            \centering
            \includegraphics[width=\linewidth]{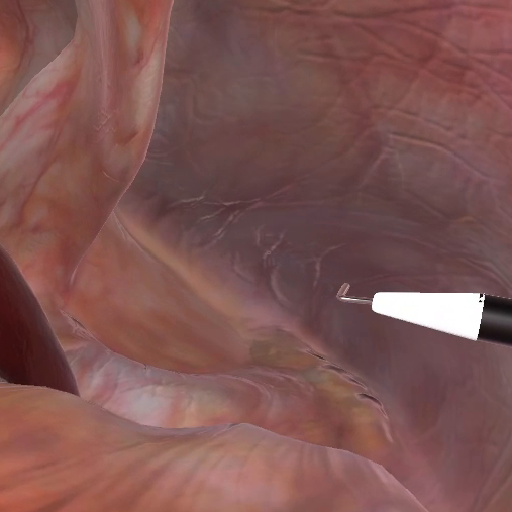}
            \caption{}
        \end{subfigure}
    \end{minipage}
    
    \begin{minipage}[c]{0.05\linewidth}
        \centering
        \rotatebox{90}{\textbf{LC-SD}}
    \end{minipage}%
    \begin{minipage}{0.95\linewidth}
        \centering
        \begin{subfigure}{0.15\linewidth}
            \centering
            \includegraphics[width=\linewidth]{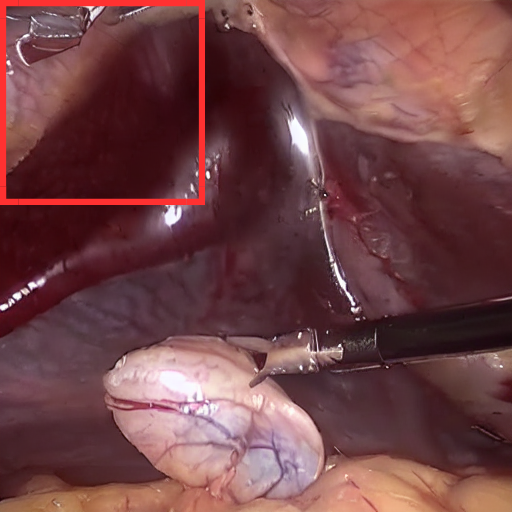}
            \caption{}
        \end{subfigure}
         \begin{subfigure}{0.15\linewidth}
            \centering
            \includegraphics[width=\linewidth]{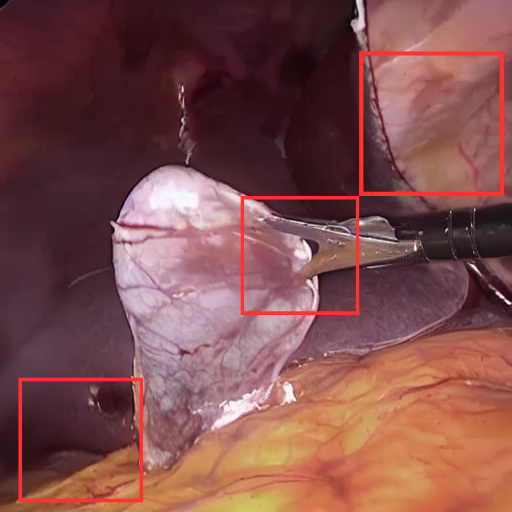}
            \caption{}
        \end{subfigure}
         \begin{subfigure}{0.15\linewidth}
            \centering
            \includegraphics[width=\linewidth]{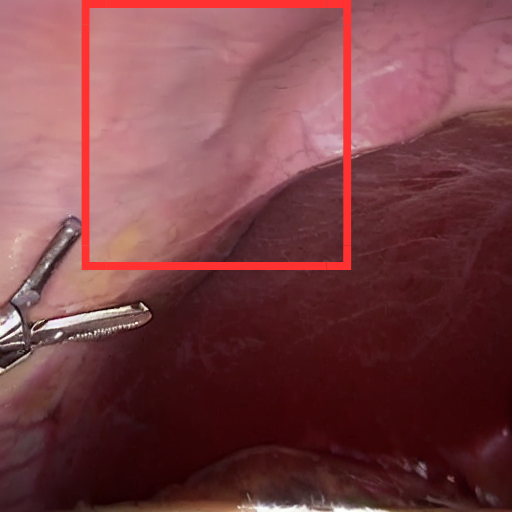}
            \caption{}
        \end{subfigure}
        \begin{subfigure}{0.15\linewidth}
            \centering
            \includegraphics[width=\linewidth]{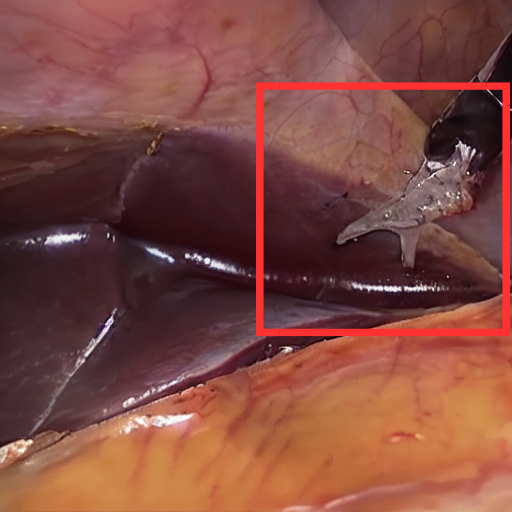}
            \caption{}
        \end{subfigure}
        \begin{subfigure}{0.15\linewidth}
            \centering
            \includegraphics[width=\linewidth]{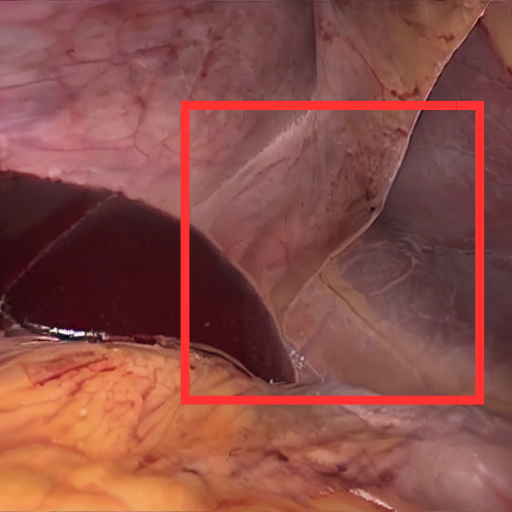}
            \caption{}
        \end{subfigure}
        \begin{subfigure}{0.15\linewidth}
            \centering
            \includegraphics[width=\linewidth]{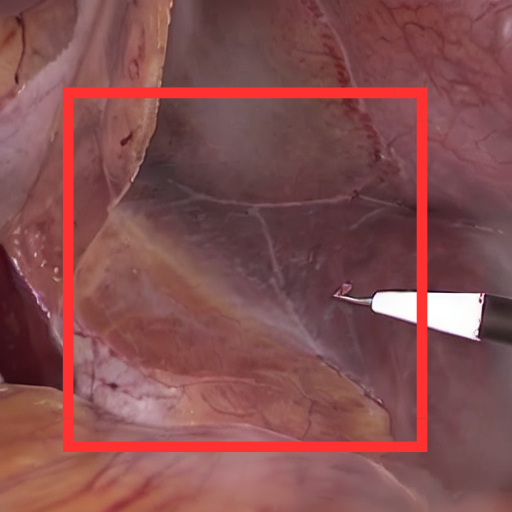}
            \caption{}
        \end{subfigure}
    \end{minipage}

    \begin{minipage}[c]{0.05\linewidth}
        \centering
        \rotatebox{90}{\textbf{\ours{}}}
    \end{minipage}%
    \begin{minipage}{0.95\linewidth}
        \centering
        \begin{subfigure}{0.15\linewidth}
            \centering
            \includegraphics[width=\linewidth]{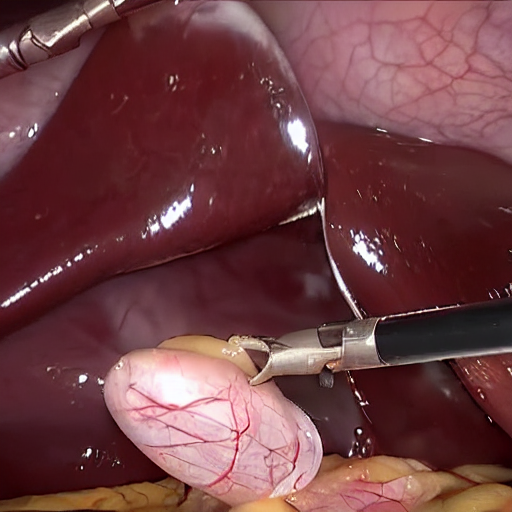}
            \caption{}
        \end{subfigure}
        \begin{subfigure}{0.15\linewidth}
            \centering
            \includegraphics[width=\linewidth]{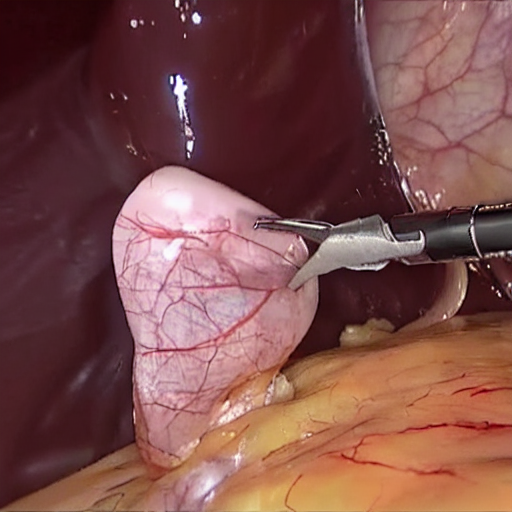}
            \caption{}
        \end{subfigure}
        \begin{subfigure}{0.15\linewidth}
            \centering
            \includegraphics[width=\linewidth]{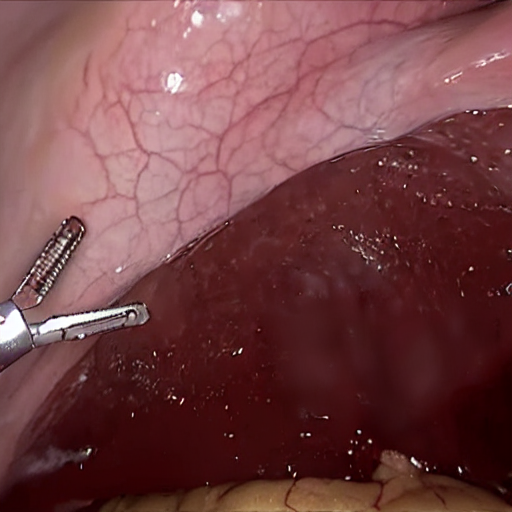}
            \caption{}
        \end{subfigure}
        \begin{subfigure}{0.15\linewidth}
            \centering
            \includegraphics[width=\linewidth]{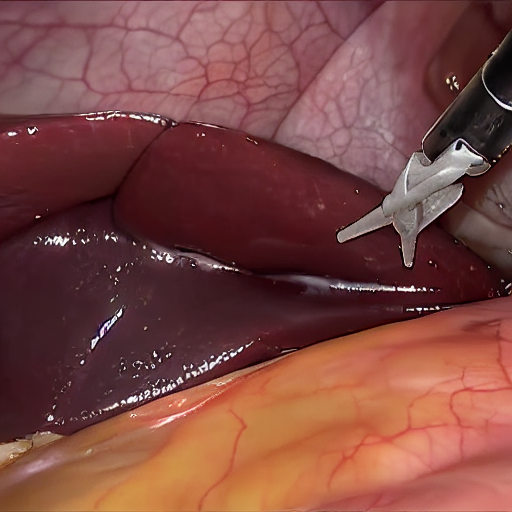}
            \caption{}
        \end{subfigure}
        \begin{subfigure}{0.15\linewidth}
            \centering
            \includegraphics[width=\linewidth]{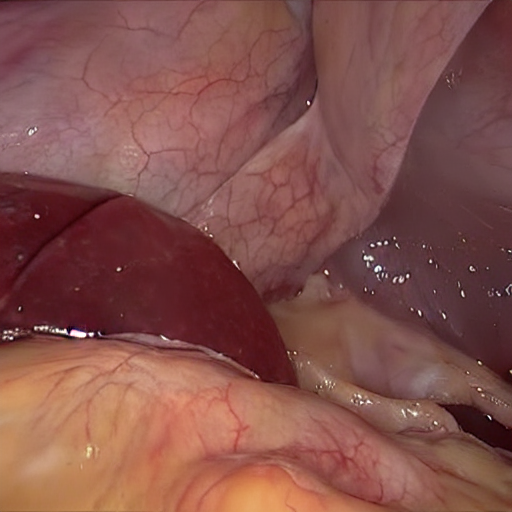}
            \caption{}
        \end{subfigure}
        \begin{subfigure}{0.15\linewidth}
            \centering
            \includegraphics[width=\linewidth]{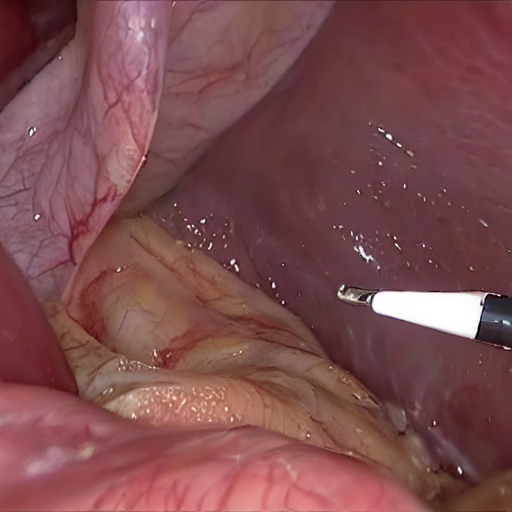}
            \caption{}
        \end{subfigure}
    \end{minipage}

    \caption{Comparison of applying LC-SD styles on our simulator to our result. Images generated using LC-SD styles appear less realistic, with blurred details. A lack of depth in the generated tissues can also be observed.}
    \label{fig:LC_SD}
\end{figure*}

For comparison, we utilized LoRAs on our simulator using the style from LC-SD, which, according to the article, yielded the best results. The visual comparison revealed that the data generated by our method achieves a similar perceptual realism to the work of LC-SD \cite{Kaleta2024} (see Figure \ref{fig:LC_SD}). Additionally, it should be noted that our simulator focuses on details and has less blurred details compared to LC-SD. We used the Mixed style, which, according to the article by \cite{Kaleta2024}, achieved the best results for comparison on our simulator. As observed, we achieved an improvement in mIoU from 67.42\% to 70.65\%. Additionally, according to the proposed metrics such as CMMD, Density, and Coverage, we observe a significant difference. These metrics highlight an increase in both fidelity and diversity of the images compared to LC-SD.
Conversely, FID and KID perform worse compared to LC-SD.

Figure \ref{fig:ControlNet} and Table \ref{tab:ablation_results} provide an overview of mIoU values for different types of control and their enhancements compared to no control inference. Table \ref{tab:ablation_results} shows that combined control models yielded the best overall results. The highest performance was achieved by using three ControlNets together. For the CholectD45 style without control, the result was 41\%. With the application of SoftEdge and Depth, the mIoU increased to 54.37\% (+13.37\%). A further increase to 58.29\% (+17.19\%) was observed with the combination of ControlNet Reference and Depth. The most significant improvement of 33.35\% was achieved with the combination of SoftEdge, Reference, and Depth control, resulting in an mIoU of 70.65\%.

\section{Discussion and Conclusions}
In this work, we introduced SimuScope, a novel framework that combines high-fidelity surgical simulation with advanced diffusion models. This innovative system facilitates the generation of photo-realistic surgical footage, which is automatically and fully labeled with essential annotations such as semantic segmentation, depth and normal maps, optical flow, action triplets, and 2D/3D instrument poses. By leveraging image-to-image translation techniques, we effectively transformed rendered images from the simulator into their realistic equivalents using SD, conditioned by multiple ControlNets.

Compared to previous works, like \cite{Kaleta2024}, which only considered a simple "fly-through" a static 3D anatomy, SimuScope supports a wide range of surgical interactions such as tissue grasping, tearing, cutting, thermo-coagulation and clipping. Such interactions are ubiquitous in surgery and containing them in the generated data is indispensable for context-aware, comprehensive and fine-grained analysis of surgical activities and workflow using deep-learning. 

Our comprehensive evaluation, employing metrics such as mIoU, FID, KID, CMMD, Density and Coverage has demonstrated that SimuScope is capable of producing photorealistic surgical images that accurately align with the simulator's output. This high degree of fidelity ensures that the synthetic data generated is both visually convincing and semantically consistent. SimuScope addresses the critical bottleneck of data scarcity by efficiently generating large datasets of high-quality imaging data. These datasets are indispensable for training and developing data-hungry deep learning models that underpin modern CAS systems.

The main limitation of this work is the lack of temporal coherency between the generated frames. As the generative AI community has recently shifted focus towards this problem (i.e., video-to-video generation), we aim to address this issue in future work.

By providing a robust and scalable solution for synthetic data generation, SimuScope holds the potential to enhance the capabilities of CAS systems significantly. This advancement can lead to safer and more precise surgical outcomes, with fewer errors and complications. Moreover, generating diverse and comprehensive datasets can accelerate research and innovation in the field, ultimately contributing to improved surgical training, better-informed clinical decisions, and enhanced patient care.

To conclude, SimuScope represents a significant step forward in integrating realistic surgical simulation with cutting-edge generative models. The success of this approach opens up new avenues for the development of sophisticated AI-driven tools in surgery, paving the way for a future where advanced CAS systems can deliver unprecedented levels of support to surgeons, thereby improving both the safety and efficacy of surgical procedures.

\section*{Acknowledgements}

This paper received funding from the European Union’s Horizon 2020 research and innovation programme under grant agreement No 857533. The research is supported by Sano project carried out within the International Research Agendas programme of the Foundation for Polish Science, co-financed by the European Union under the European Regional Development Fund. The research was created within the project of the Minister of Science and Higher Education ”Support for the activity of Centers of Excellence established in Poland under Horizon 2020” on the basis of the contract number MEiN/2023/DIR/3796. We gratefully acknowledge Polish high-performance computing infrastructure PLGrid (HPC Centers: ACK Cyfronet AGH) for providing computer facilities and support within computational grant no. PLG/2024/017108.

{\small
\bibliographystyle{ieee_fullname}
\bibliography{egbib}
}

\newpage

\setcounter{figure}{0}                      %
\setcounter{table}{0}                       %

\twocolumn[
    \begin{center}
        \section*{-- Supplementary Material --}
        \vspace{0.5em}
        \Large
        SimuScope: Realistic Endoscopic Synthetic Dataset Generation through Surgical Simulation and Diffusion Models
        \vspace{1em}
    \end{center}
]

\section*{A. Publicly available dataset}
\begin{figure*}[h!]
    \centering
    \captionsetup[subfigure]{labelformat=empty}
    \begin{subfigure}{0.23\linewidth}
        \centering
        \includegraphics[width=\linewidth]{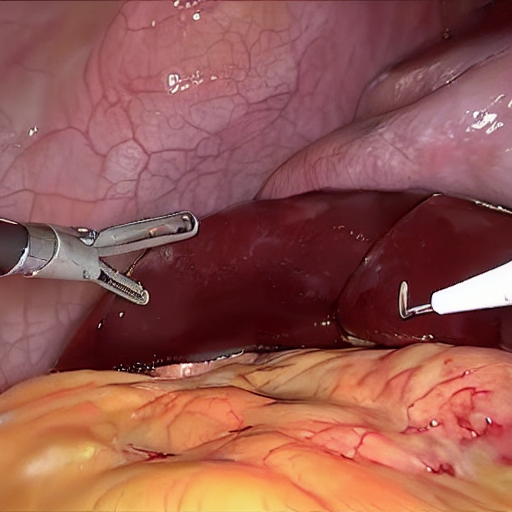}
        \caption{}
    \end{subfigure}
    \begin{subfigure}{0.23\linewidth}
        \centering
        \includegraphics[width=\linewidth]{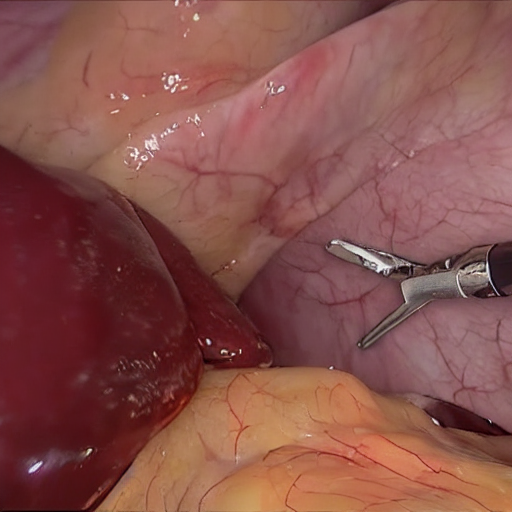}
        \caption{}
    \end{subfigure}
    \begin{subfigure}{0.23\linewidth}
        \centering
        \includegraphics[width=\linewidth]{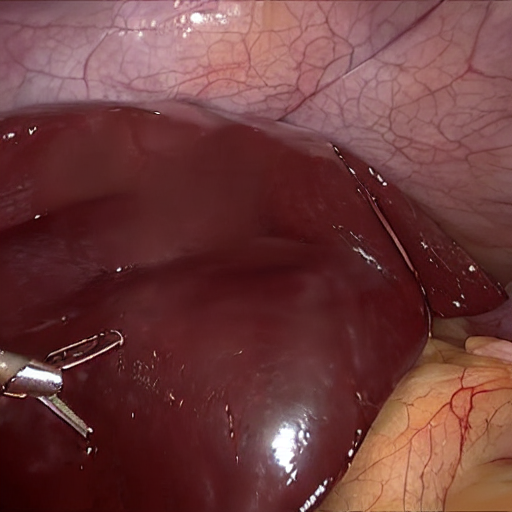}
        \caption{}
    \end{subfigure}
    \begin{subfigure}{0.23\linewidth}
        \centering
        \includegraphics[width=\linewidth]{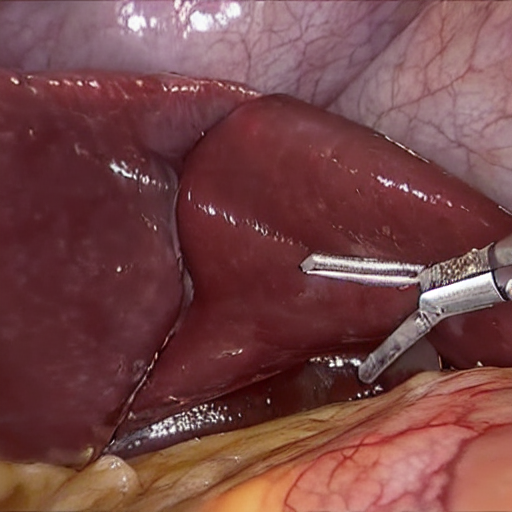}
        \caption{}
    \end{subfigure}
    \vspace{0.3cm}
    \begin{subfigure}{0.23\linewidth}
        \centering
        \includegraphics[width=\linewidth]{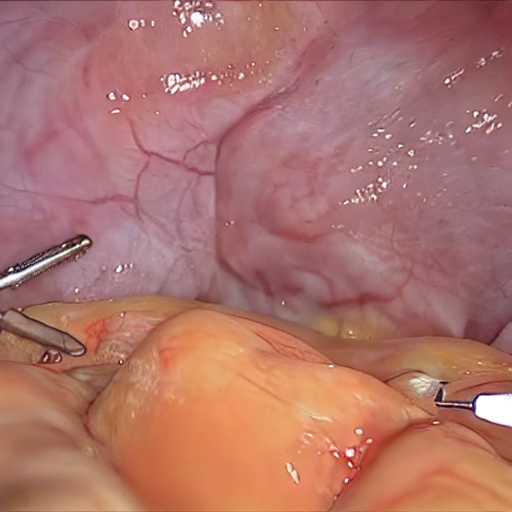}
        \caption{}
    \end{subfigure}
    \begin{subfigure}{0.23\linewidth}
        \centering
        \includegraphics[width=\linewidth]{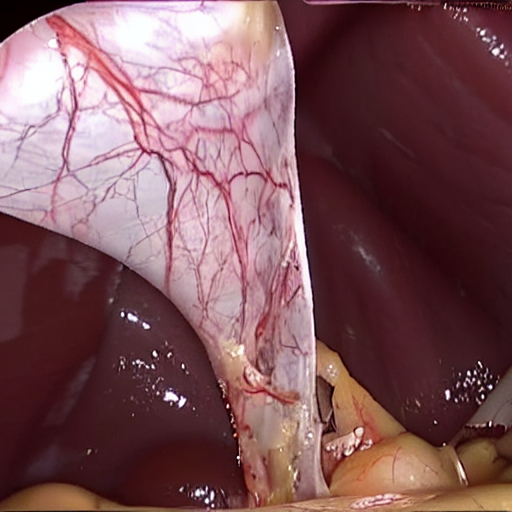}
        \caption{}
    \end{subfigure}
    \begin{subfigure}{0.23\linewidth}
        \centering
        \includegraphics[width=\linewidth]{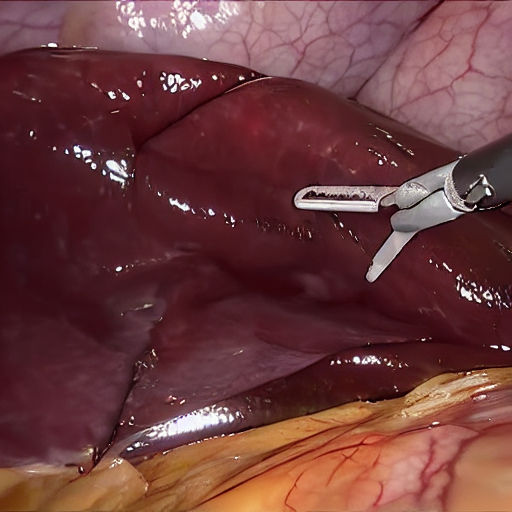}
        \caption{}
    \end{subfigure}
    \begin{subfigure}{0.23\linewidth}
        \centering
        \includegraphics[width=\linewidth]{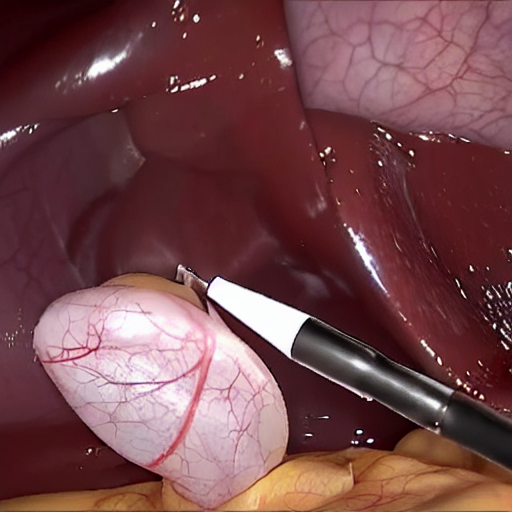}
        \caption{}
    \end{subfigure}
    \vspace{0.3cm}
    \begin{subfigure}{0.23\linewidth}
        \centering
        \includegraphics[width=\linewidth]{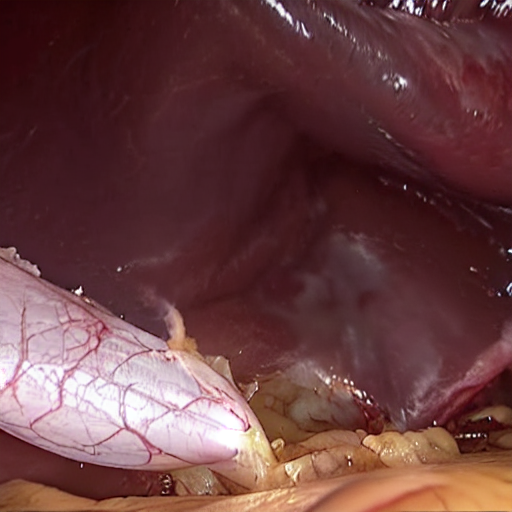}
        \caption{}
    \end{subfigure}
    \begin{subfigure}{0.23\linewidth}
        \centering
        \includegraphics[width=\linewidth]{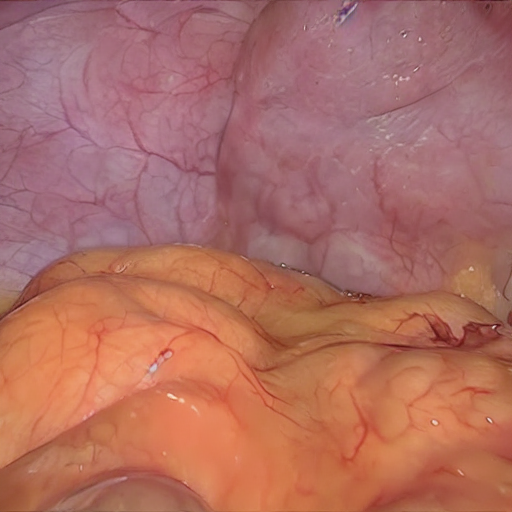}
        \caption{}
    \end{subfigure}
    \begin{subfigure}{0.23\linewidth}
        \centering
        \includegraphics[width=\linewidth]{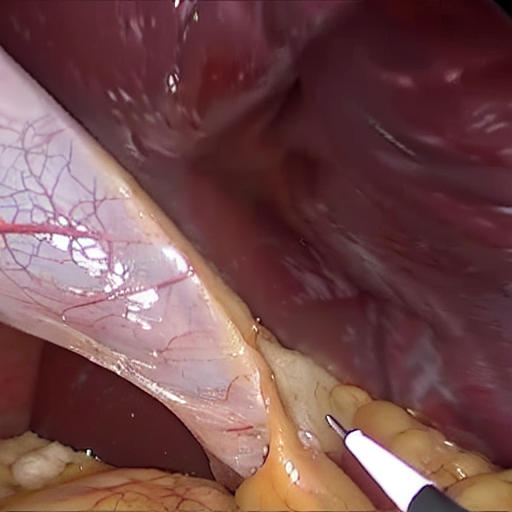}
        \caption{}
    \end{subfigure}
    \begin{subfigure}{0.23\linewidth}
        \centering
        \includegraphics[width=\linewidth]{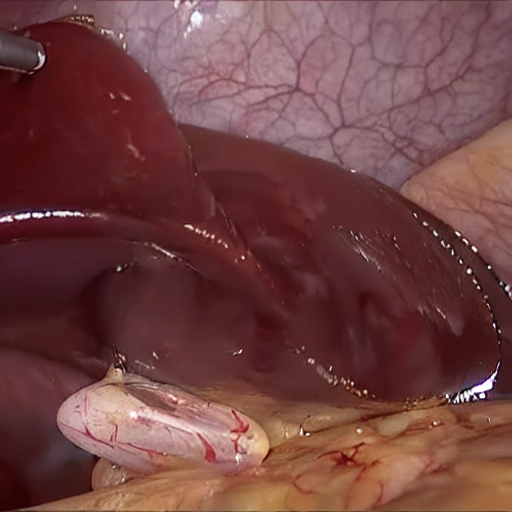}
        \caption{}
    \end{subfigure}
    \vspace{0.3cm}
    \begin{subfigure}{0.23\linewidth}
        \centering
        \includegraphics[width=\linewidth]{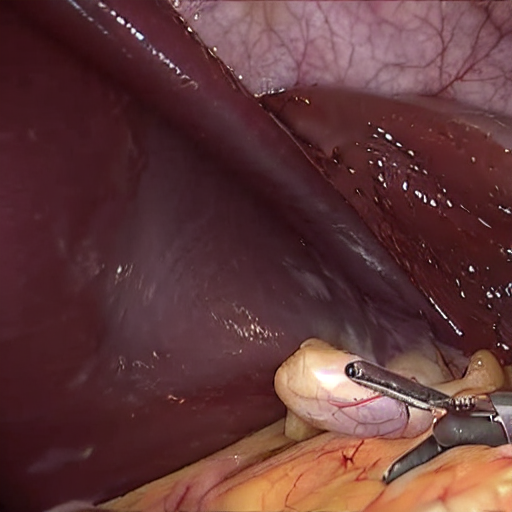}
        \caption{}
    \end{subfigure}
    \begin{subfigure}{0.23\linewidth}
        \centering
        \includegraphics[width=\linewidth]{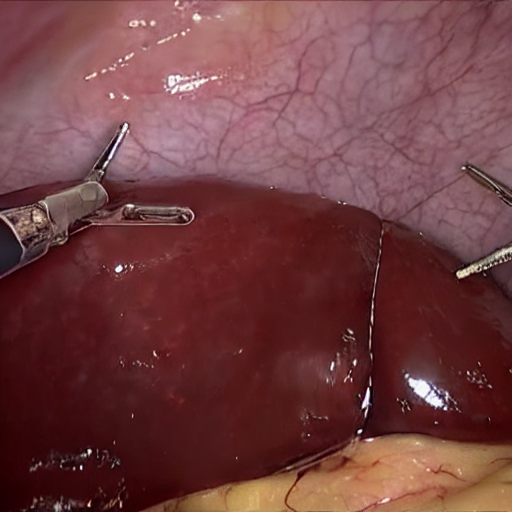}
        \caption{}
    \end{subfigure}
    \begin{subfigure}{0.23\linewidth}
        \centering
        \includegraphics[width=\linewidth]{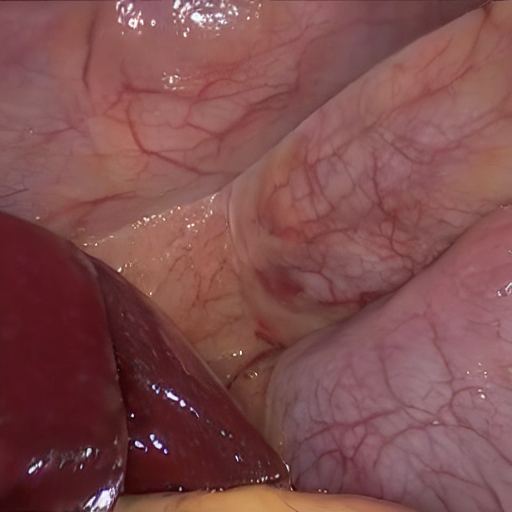}
        \caption{}
    \end{subfigure}
    \begin{subfigure}{0.23\linewidth}
        \centering
        \includegraphics[width=\linewidth]{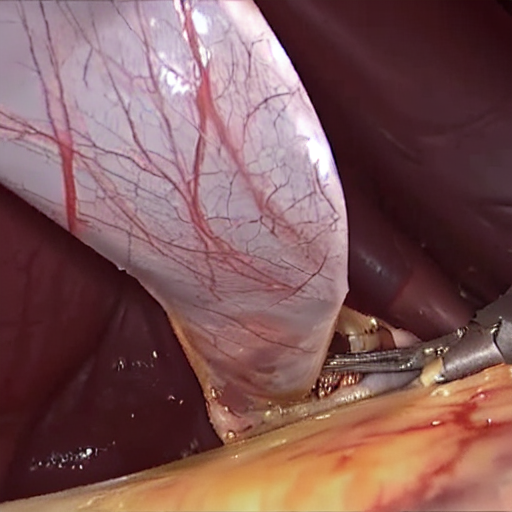}
        \caption{}
    \end{subfigure}
    \caption{SimuScope generated images. A collection of sample generated images from our simulator is presented here, showcasing a diverse range of perspectives. These examples highlight the simulator's capability to produce detailed visualizations from different angles and orientations. The images encompass various anatomical regions and demonstrate the versatility of the simulation in replicating realistic scenarios.}
    \label{fig:ADD}
\end{figure*}

\begin{figure*}[h!]
    \centering
    \captionsetup[subfigure]{labelformat=empty}
    \begin{subfigure}{0.23\linewidth}
        \centering
        \includegraphics[width=\linewidth]{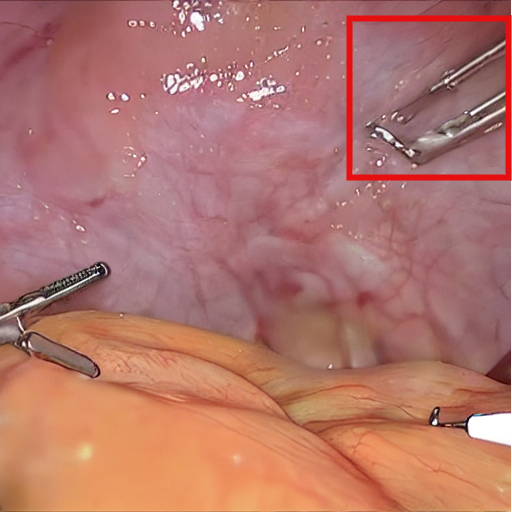}
        \caption{}
    \end{subfigure}
    \begin{subfigure}{0.23\linewidth}
        \centering
        \includegraphics[width=\linewidth]{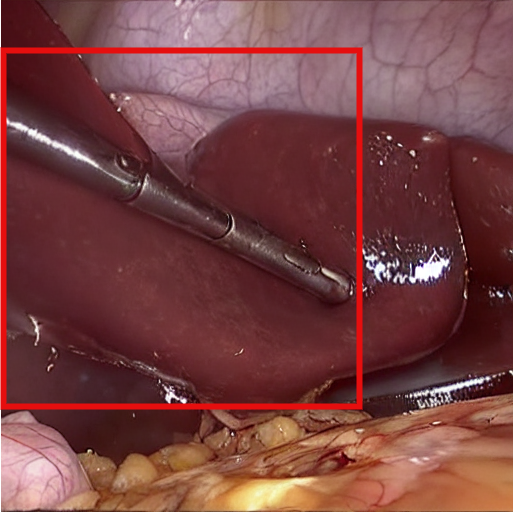}
        \caption{}
    \end{subfigure}
    \begin{subfigure}{0.23\linewidth}
        \centering
        \includegraphics[width=\linewidth]{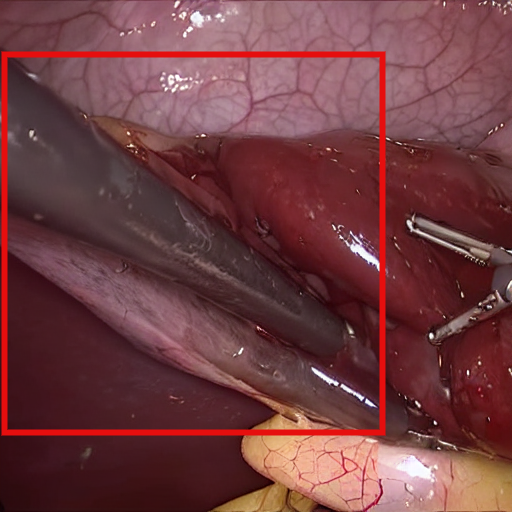}
        \caption{}
    \end{subfigure}
    \begin{subfigure}{0.23\linewidth}
        \centering
        \includegraphics[width=\linewidth]{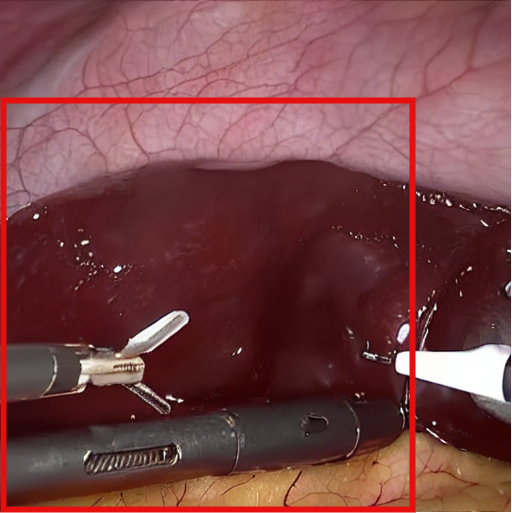}
        \caption{}
    \end{subfigure}
       \begin{subfigure}{0.23\linewidth}
        \centering
        \includegraphics[width=\linewidth]{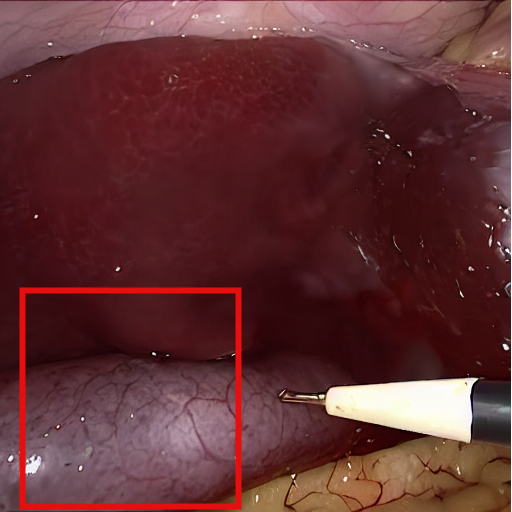}
        \caption{}
    \end{subfigure}
    \begin{subfigure}{0.23\linewidth}
        \centering
        \includegraphics[width=\linewidth]{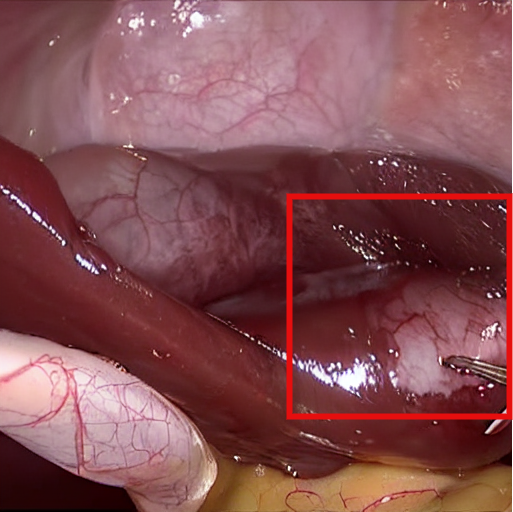}
        \caption{}
    \end{subfigure}
    \begin{subfigure}{0.23\linewidth}
        \centering
        \includegraphics[width=\linewidth]{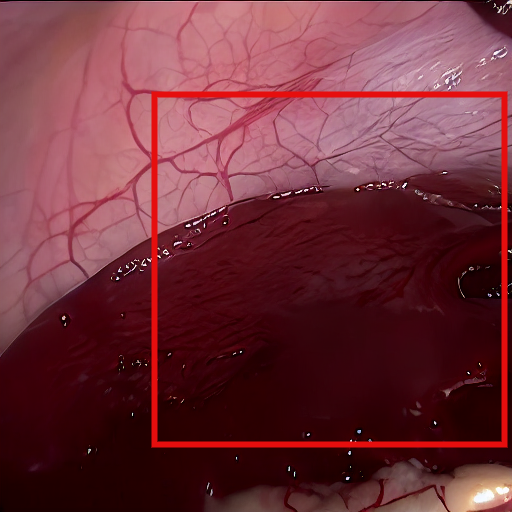}
        \caption{}
    \end{subfigure}
    \begin{subfigure}{0.23\linewidth}
        \centering
        \includegraphics[width=\linewidth]{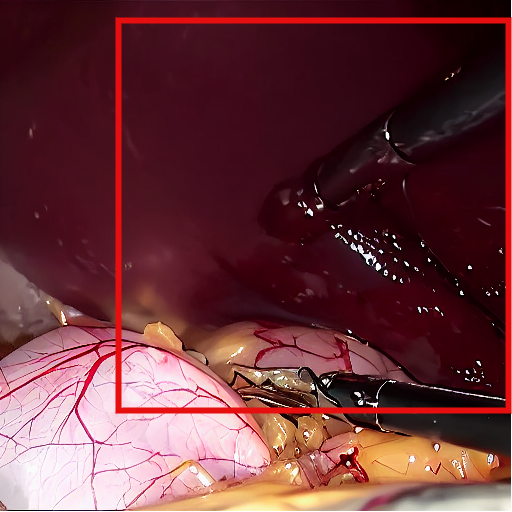}
        \caption{}
    \end{subfigure}
    \caption{SimuScope artifacts. During the generation of images, various artifacts appeared. As shown in the attached images, these mainly include the addition of instruments in parts of the liver or abdominal wall, which can be observed in the first four images from the first row. These artifacts may interfere with accurate interpretation by introducing extraneous elements that do not belong to the actual anatomical structures. The next four images from the second row exhibit artifacts related to color saturation, where abnormal intensities and hues may distort the visual information. These color saturation artifacts can obscure important details and mislead diagnostic assessments by creating false impressions of tissue characteristics.}
    \label{fig:Artifacts}
\end{figure*}

To foster research in the field, we release our dataset for a community. Here, we present more generated surgical images provided by our proposed method called SimuScope. In total, the simulator dataset consists of 13,064 images. During the generation of realistic images (see Figure \ref{fig:ADD}), artifacts appeared on the generated images (see Figure \ref{fig:Artifacts}) which have also been attached. The number of images with artifacts is 1673 out of 13064.

Figure \ref{fig:ADD} presents a wide range of sample images generated by SimuScope, showcasing various perspectives and anatomical regions during cholecystectomy. These examples emphasize the simulator's capability to produce high-fidelity visualizations from different angles, providing a detailed representation of the surgical environment.

Figure \ref{fig:Artifacts} presents artifacts that emerged during the SimuScope image generation process, which are a natural consequence of the simulator's high realism. The complex lighting effects and diverse textures used in the simulator contribute to these artifacts, reflecting the intricate nature of realistic surgical environments. While some of the artifacts, such as instrument placement or color saturation inconsistencies, may appear, they result from the simulator's effort to accurately replicate real-world scenarios, highlighting areas for further refinement in the simulation.

\begin{figure*}[t!]
    \centering
    \vspace*{0.2cm}
    \begin{subfigure}{0.23\linewidth}
        \includegraphics[width=\linewidth]{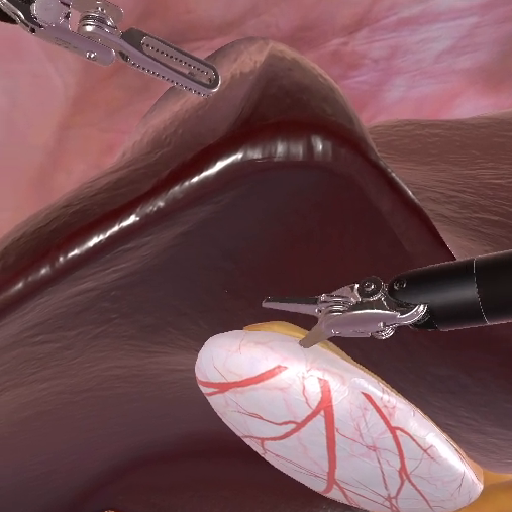}
        \caption{}
    \end{subfigure}
    \begin{subfigure}{0.23\linewidth}
        \includegraphics[width=\linewidth]{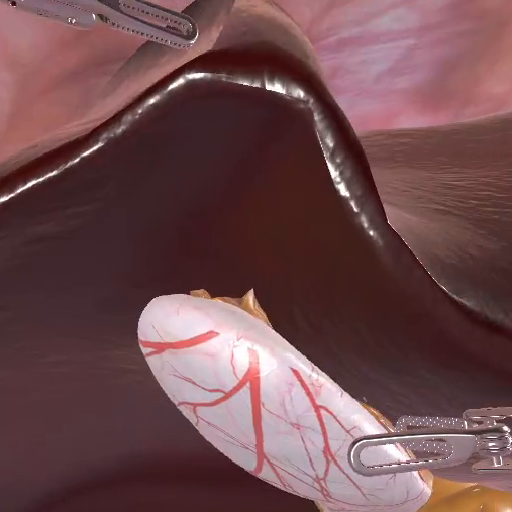}
        \caption{}
    \end{subfigure}
    
    \begin{subfigure}{0.23\linewidth}
        \includegraphics[width=\linewidth]{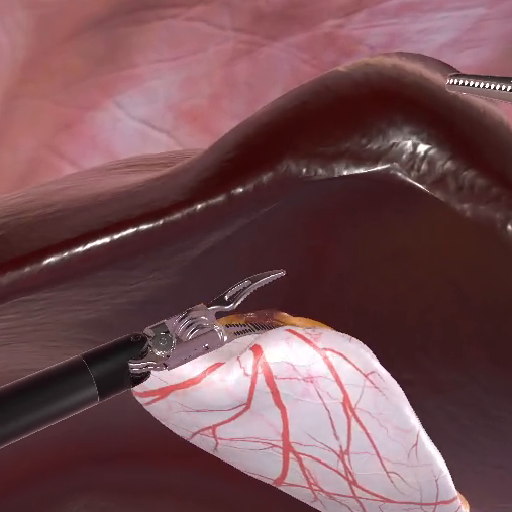}
        \caption{}
    \end{subfigure}
    \begin{subfigure}{0.23\linewidth}
        \includegraphics[width=\linewidth]{figures/Maryland_bipolar_forceps.png}
        \caption{}
    \end{subfigure}
    \begin{subfigure}{0.23\linewidth}
        \includegraphics[width=\linewidth]{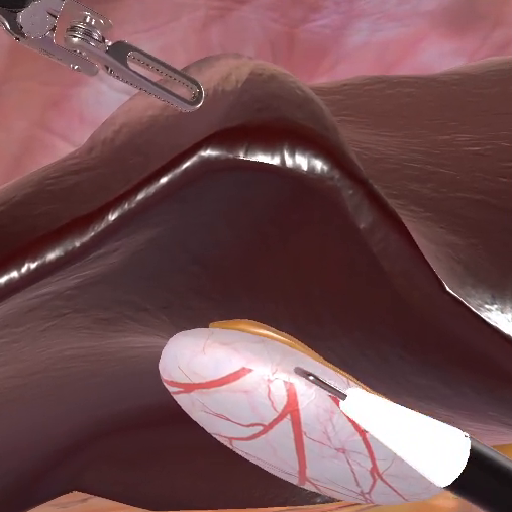}
        \caption{}
    \end{subfigure}
   
  \caption{Images showing DaVinci tools used in the cholecystectomy simulator: a) Round Tip Scissors - versatile scissors for delicate tissue manipulation, providing precise cuts and control during the procedure, b) Cadiere Forceps - essential for grasping and holding tissues securely, allowing for firm yet gentle manipulation of tissues, c) Maryland Bipolar Forceps - used for precise tissue dissection and coagulation, offering both cutting and hemostatic capabilities, d) Monopolar Curved Scissors - precision tool for cutting and dissecting tissues, designed to facilitate curved and angled incisions, and e) Hook - versatile instrument for hooking and retracting tissues, aiding in the exposure and manipulation of the surgical site.}

    \label{fig:DaVinci}
\end{figure*}

Figure \ref{fig:DaVinci} illustrates the Da Vinci surgical tools in the dataset used for the cholecystectomy simulator. The dataset includes a diverse range of robotic instruments, enabling realistic modeling of surgical procedures. These tools are essential for replicating the precise and controlled actions of robotic-assisted surgery, making the dataset a valuable resource for developing and evaluating simulation-based methodologies. The variety of instruments allows the simulator to closely mirror real-world surgical techniques, enhancing its utility for both training and research applications.

\section*{B. Visual evaluation and parameter configuration for LoRA adapters}
Table \ref{tab:parameter} presents the selected experimental parameter values for various LoRA (Low-Rank Adaptation) configurations applied to the models. It highlights key parameters such as Denoising Strength, CFG (Classifier-Free Guidance), SoftEdge Strength, Depth Control Strength, and Reference Control Strength. All experiments consistently utilize the DPM++2M Karras noise scheduler to maintain uniform noise handling across simulations, thereby improving the quality and reliability of the generated outputs. For the specific case of the CholectD45 model, the parameter values are: Denois at 0.65, CFG at 7.0, SoftEdge at 0.45, Depth at 0.65, and Reference at 0.65.

\begin{figure*}[h!]
    \centering
    \captionsetup[subfigure]{labelformat=empty}
    \begin{subfigure}{0.23\linewidth}
        \centering
        \includegraphics[width=0.9\linewidth]{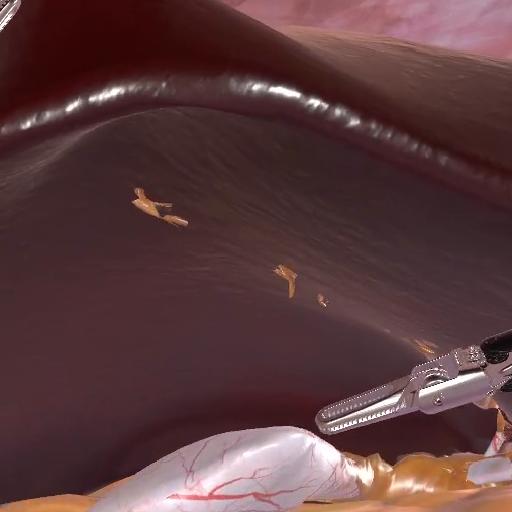}
        \caption{Simulator}
    \end{subfigure}
    \begin{subfigure}{0.23\linewidth}
        \centering
        \includegraphics[width=0.9\linewidth]{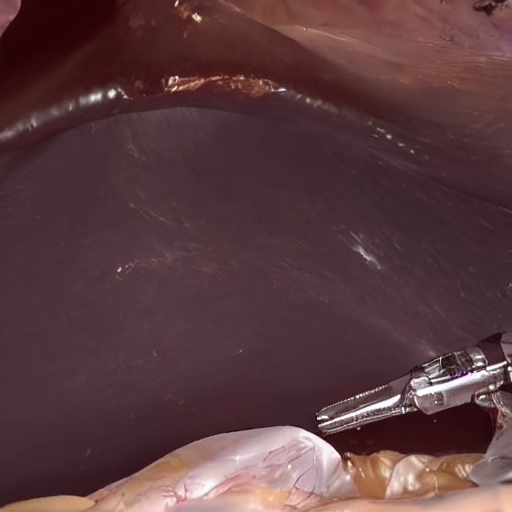}
        \caption{CholectG45 [0.45]}
    \end{subfigure}
    \begin{subfigure}{0.23\linewidth}
        \centering
        \includegraphics[width=0.9\linewidth]{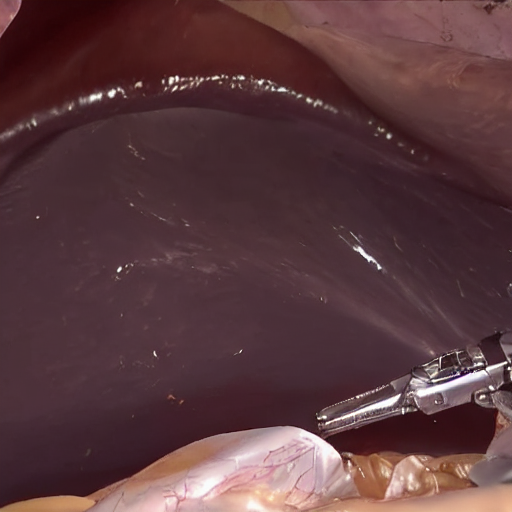}
        \caption{CholectL45 [0.45]}
    \end{subfigure}
    \begin{subfigure}{0.23\linewidth}
        \centering
        \includegraphics[width=0.9\linewidth]{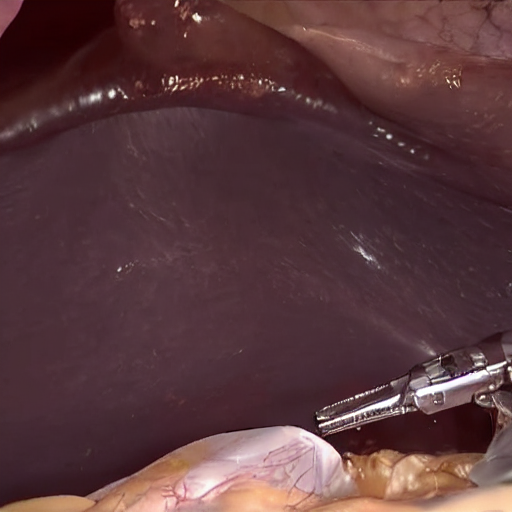}
        \caption{CholectG45 [0.6]}
    \end{subfigure}

    \vspace{0.3cm}

    \begin{subfigure}{0.23\linewidth}
        \centering
        \includegraphics[width=0.9\linewidth]{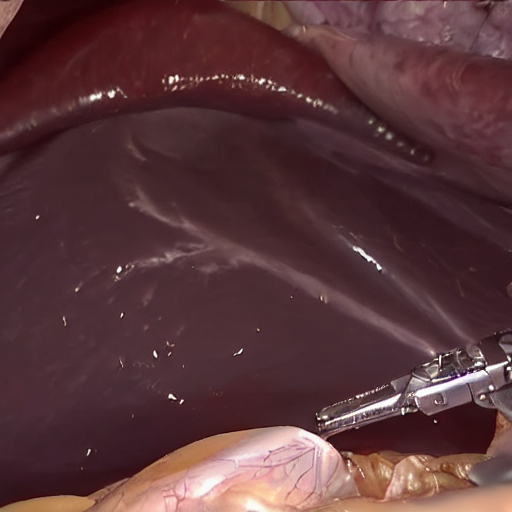}
        \caption{CholectL45 [0.6]}
    \end{subfigure}
    \begin{subfigure}{0.23\linewidth}
        \centering
        \includegraphics[width=0.9\linewidth]{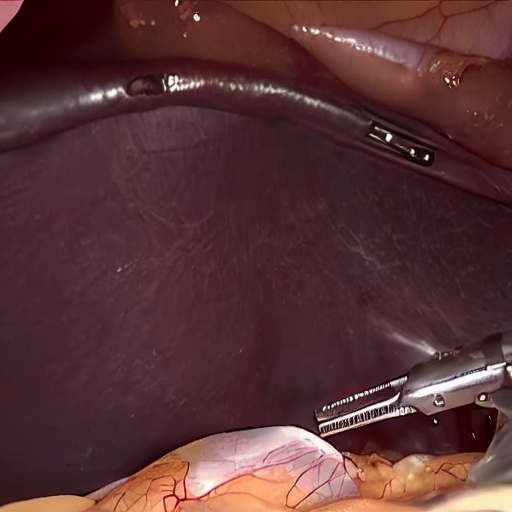}
        \caption{CholectG45 [0.9]}
    \end{subfigure}
    \begin{subfigure}{0.23\linewidth}
        \centering
        \includegraphics[width=0.9\linewidth]{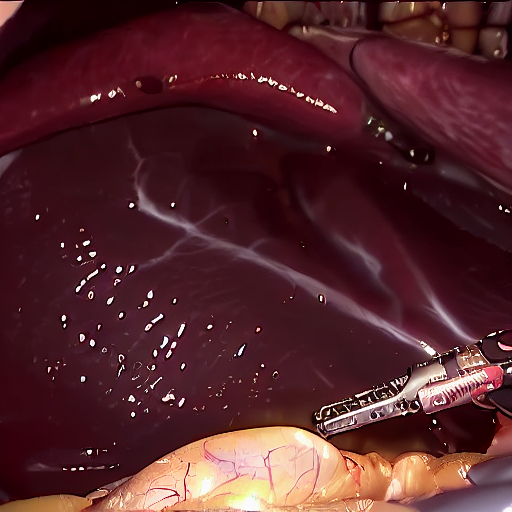}
        \caption{CholectL45 [0.9]}
    \end{subfigure}
    \begin{subfigure}{0.23\linewidth}
        \centering
        \includegraphics[width=0.9\linewidth]{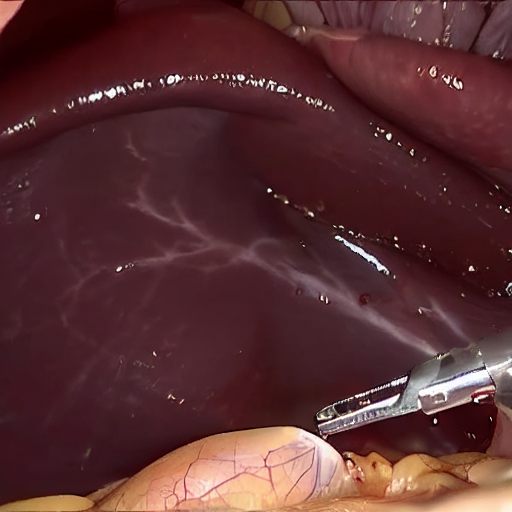}
        \caption{CholectD45 [0.45]}
    \end{subfigure}

    \caption{Visual comparisons reveal that using LoRAs individually degrades results. At 0.45 weight, CholectD45 shows minimal improvement. At 0.6, tissue and instrument appearance improves, though results remain synthetic. At 0.9, color artifacts appear. Combining two LoRAs yields the best results, closely resembling real endoscopic images with enhanced realism and fidelity.}
    \label{fig:Lora}
\end{figure*}

This set of images (see Figure \ref{fig:Lora}) illustrates the outcomes of experiments using different LoRA adapter with varying parameter settings. The figure shows a series of visual comparisons at different LoRA weights (0.45, 0.6, and 0.9) across multiple configurations, highlighting how the integration of LoRAs affects the realism of simulated endoscopic images. At lower weights, the results are less realistic, with visible artifacts and a synthetic appearance. At higher weights, improvements in tissue and instrument rendering are observed, although color artifacts appear. The combination of multiple LoRAs provides the most realistic and faithful simulation.


\begin{table}[H]
\caption{Selected experimentaly LoRA's parameter values for each model: denoising strength, CFG, SoftEdge, Depth, and Reference control strength. All models utilize the noise scheduler DPM++2M Karras, ensuring consistent noise management across simulations and enhancing the fidelity of generated outputs.}
\centering
\resizebox{0.5\textwidth}{!}{%
\begin{tabular}{l c c c c c c}
\toprule
Style & Denois & CFG & SoftEdge & Depth & Reference \\ 
\midrule

 CholectD45 & 0.65 & 7.0 & 0.45 & 0.65 & 0.65 \\
\bottomrule
\end{tabular}
}
\label{tab:parameter}
\end{table}

\end{document}